\documentclass[preprint,3p,11pt]{elsarticle}

\usepackage{amssymb}
\usepackage{amsthm}

\usepackage[caption=false,font=footnotesize,labelfont=sf,textfont=sf]{subfig}
\usepackage{booktabs}
\usepackage[flushleft]{threeparttable}
\usepackage{setspace}
\usepackage[algoruled, longend, linesnumbered, noline]{algorithm2e}
\usepackage[plainpages=false, bookmarks=false]{hyperref}           % para referência cruzadas
\hypersetup{
    colorlinks = true,
    linkcolor = blue,
    citecolor = blue,
    urlcolor = blue
}
\usepackage{longtable}  

\newtheorem{definition}{Definition}

\journal{Information Sciences}

\begin{document}

\begin{frontmatter}


% \author{Name\corref{cor1}\fnref{label2}}
% \address{Address\fnref{label3}}

\title{Time Series Clustering via Community Detection in Networks\tnoteref{t1}}


\tnotetext[t1]{This manuscript version is made available under the \href{http://creativecommons.org/licenses/by-nc-nd/4.0/}{Creative Commons BY-NC-ND 4.0 license}.}

\author[rvt]{Leonardo N. Ferreira\corref{cor1}}
\ead{leonardo@icmc.usp.br}

\author[focal]{Liang Zhao}
\ead{zhao@usp.br}

\cortext[cor1]{Corresponding author}

\address[rvt]{Institute of Mathematics and Computer Science, University of S\~ao Paulo\\Av. Trabalhador S\~ao-carlense, 400 CEP: 13566-590 - Centro, S\~ao Carlos - SP, Brazil.}
\address[focal]{Department of Computing and Mathematics, University of S\~ao Paulo\\Av. Bandeirantes, 3900 - CEP: 14040-901 - Monte Alegre - Ribeir\~ao Preto - SP, Brazil.}

% \author{Leonardo N. Ferreira\corref{cor1}}
% \ead{leonardo@icmc.usp.br}
% \address{Institute of Mathematics and Computer Science, University of S\~ao Paulo\\S\~ao Carlos - SP, Brazil.}

% \author{Liang Zhao\fnref{fn2}}
% \ead{zhao@usp.br}
% \address{Department of Computing and Mathematics, University of S\~ao Paulo\\Ribeir\~ao Preto - SP, Brazil.}

% \cortext[cor1]{Corresponding author}

\begin{abstract}
In this paper, we propose a technique for time series clustering using community detection in complex networks. Firstly, we present a method to transform a set of time series into a network using different distance functions, where each time series is represented by a vertex and the most similar ones are connected. Then, we apply community detection algorithms to identify groups of strongly connected vertices (called a community) and, consequently, identify time series clusters. Still in this paper, we make a comprehensive analysis on the influence of various combinations of time series distance functions, network generation methods and community detection techniques on clustering results. Experimental study shows that the proposed network-based approach achieves better results than various classic or up-to-date clustering techniques under consideration. Statistical tests confirm that the proposed method outperforms some classic clustering algorithms, such as $k$-medoids, diana, median-linkage and centroid-linkage in various data sets. Interestingly, the proposed method can effectively detect shape patterns presented in time series due to the topological structure of the underlying network constructed in the clustering process. At the same time, other techniques fail to identify such patterns. Moreover, the proposed method is robust enough to group time series presenting similar pattern but with time shifts and/or amplitude variations. In summary, the main point of the proposed method is the transformation of time series from time-space domain to topological domain. Therefore, we hope that our approach contributes not only for time series clustering, but also for general time series analysis tasks.
\end{abstract}


\end{frontmatter}


\section{Introduction}
\label{sec:introduction}

Time series data mining has received a lot of attention in the last years due to the ubiquity of this kind of data. One specific task is clustering with the goal to divide a set of time series into groups, where similar ones are put in the same cluster \cite{esling12}. Such kind of problem has been observed in many application domains like climatology, geology, health sciences, energy consumption, failure detection, among others \cite{liao05}.

The two desired aspects when performing time series clustering are effectiveness and efficiency \cite{keogh13}. Effectiveness can be achieved by representation methods that should be capable of dealing with high dimensional data. Efficiency is obtained by using distance functions and clustering algorithms that can properly distinguish different time series in an efficient way. Keeping these two features in mind, many clustering algorithms have been proposed and those can be broadly classified into two approaches: data-adaptation and algorithm-adaptation \cite{liao05}. The former extracts features arrays from each time series and then applies a clustering algorithm in its original form. The latter uses specially designed clustering algorithms to directly handle time series. In this case, the major modification is the distance function, which should be capable of distinguishing time series. 

Complex networks form a recent and interesting research area. Here, a complex network refers to a large scale network with non trivial connection pattern \cite{boccaletti06}. Many real-world systems can be modeled by networks. One of the salient features in many networks is the presence of community structure, which is represented by groups of densely connected vertices and, at the same time, with sparse connections between groups. Detecting such structures is interesting in many real applications. For this reason, many community detection algorithms have been developed \cite{fortunato10} and such algorithms present a powerful mechanism for general data mining tasks. A brief review of community detection techniques will be given in the next section.  

In the original form of time series, only the local relationship among neighbor data samples can be easily identified, while long distance global relationship remains unknown in general. On the other hand, time series analysis, such as time series clustering, classification or prediction, requires not only local information, but also global knowledge to capture the pattern formation of a given time series. Network (graph) is a powerful mechanism, which is able to characterize the relationship between any pair or any groups of data samples. Therefore, the transformation from time series to network representation is hopefully to present an alternative way for time series analysis. From the technical view point, network-based clustering techniques also present attractive advantage. Up to now, the majority of existing time series clustering techniques in literature use \emph{k}-means, \emph{k}-medoids or hierarchical clustering algorithms in their original forms or modified versions. The common feature of these algorithms is that they try to break data samples into clusters in such a way that the partition optimizes a criterion defined by a given distance function. As a consequence,  these techniques can just find clusters of a specific shape already determined by the distance function. For example, \emph{k}-means with the Euclidean distance function can only produce Gaussian distributed clusters. On the other hand, it has been shown that network-based clustering techniques can capture arbitrary cluster shapes. This is because network-based techniques identify connectivity patterns of the input data and such patterns can be any shape in the Euclidean space. Finally, many community detection techniques have been proposed and some of them have even linear time complexity when the constructed network is sparse \cite{SilvaZhao2012}. This feature also makes them attractive to time series data clustering.

In this paper, we aim to apply network science to temporal data mining. We intend to verify the benefits of using community detection algorithms in time series data clustering. More specifically, we propose an algorithm including 4 steps of processing: (1) data normalization; (2) distance function calculation; (3) network construction, where every vertex represents a time series connected to its most similar ones using a distance function; (4) community detection, where each community represents a time series cluster. 
In summary, this paper presents the following contributions:

\begin{itemize}
  \item The main contribution is the proposal of using community detection in complex networks for time series clustering. For this purpose, we transform time series from time-space domain to topological domain. Since network is a general representation, which has ability to characterize both local and global relationship among nodes (representing data samples), therefore, our approach is useful not only for time series clustering but also for other kinds of time series analysis tasks. To our knowledge, applying community detection techniques for time series clustering has not been reported in the literature;
  \item Extensive numerical study has been conducted in this paper. Specifically, we study, in the time series clustering context, combinations of time series data sets, time series distance functions, network construction methods and community detection algorithms. In comparison to other time series clustering algorithms, experimental results and statistical tests show that the network-based approach present better results.
  \item Last but not least, the proposed method presents some desired features when applied to real clustering problems. It can effectively detect shape patterns presented in time series due to the topological structure of the underlying network constructed in the clustering process. At the same time, other techniques studied in this paper fail to identify such patterns. Moreover, the proposed method is robust enough to group time series presenting similar pattern but with time shifts and/or amplitude variations.
\end{itemize}

The remainder of this paper is organized as follows. Firstly, we present in Section \ref{sec:background} some background concepts and related works to this paper. In Sections \ref{sec:method} and \ref{sec:experiments} we present our approach and the experimental results, respectively. Finally, we point some final remarks and future works in Section \ref{sec:final}.

% ==============================================================================
% Section -
% ==============================================================================
\section{Background and related works}
\label{sec:background}

In this section, we review the three main components of time series clustering used in this paper: time series distance measures, clustering algorithms and community detection in networks.

\subsection{Time series distance measures}

We start by presenting the basic concept: time series. For simplicity and without loss of generality, we assume that time is discrete.

\begin{definition}[Time Series]
    A time series $X$ is an ordered sequence of $t$ real values $ X = \{ x_1, \dots , x_t\}, x_i \in \mathbb{R},i \in \mathbb{N}$.
\end{definition}

The main idea of clustering is to group similar objects. In order to discover which data are similar, several distance (or dissimilarity) measures were defined in the literature. In this paper, we use the terms ``similarity'' and ``distance'' in inverse concepts. In the case of time series distance measures, the distance measures can be classified into four categories \cite{esling12}: shape-based, edit-based, feature-based, and structure-based.

\subsubsection{Shaped based distance measures} 

\begin{figure}[ht]
    \centering
    \subfloat[]{\label{fig:align1} \includegraphics[scale=0.28, trim = 10mm 15mm 10mm 15mm]{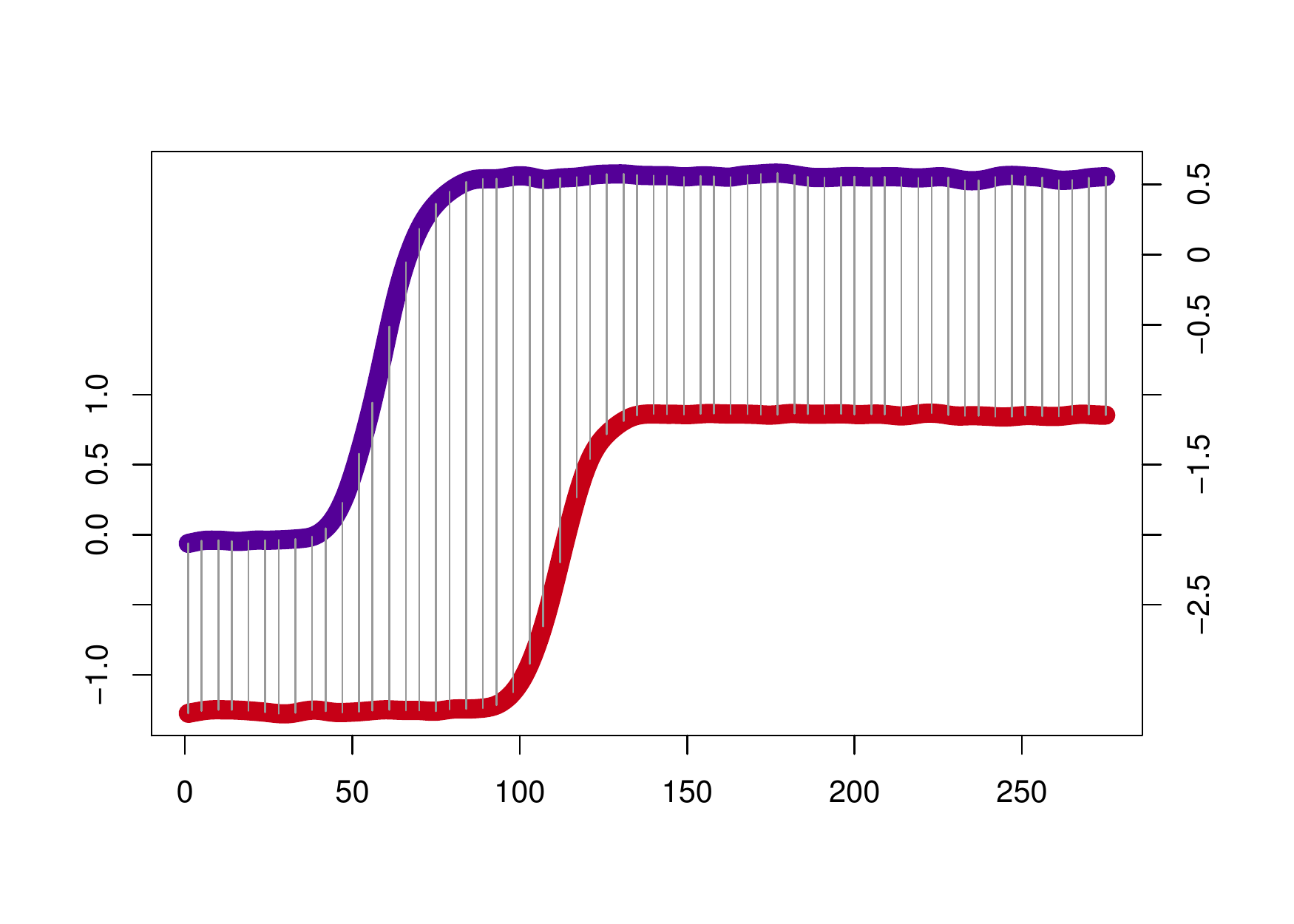}} 
    \subfloat[]{\label{fig:align2} \includegraphics[scale=0.28, trim = 10mm 15mm 10mm 15mm]{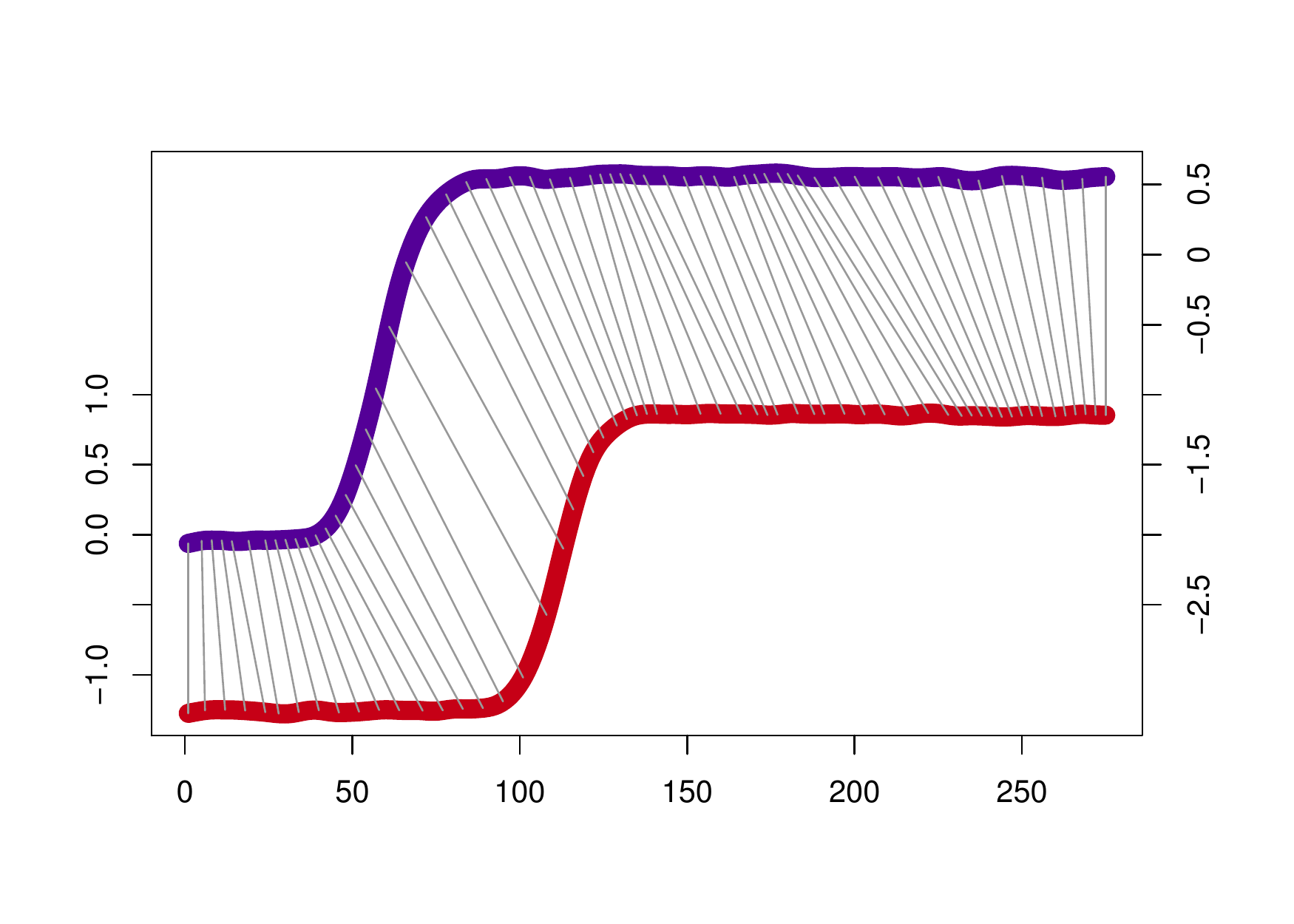}} 
    \caption{Time series comparison using lock-step and elastic measures, respectively. The total distance is proportional to the length of the gray lines. \protect\subref{fig:align1} Lock-step measures compare fixed (one-to-one) pairs of elements. \protect\subref{fig:align2} Elastic measures perform the alignment of the series and allow one-to-many comparisons of elements.}
    \label{fig:align} 
   \end{figure}

The first category of time series distance measures is based on the shape of the time series. Such measures compare directly the raw data of a pair of time series. The most common measures are the $L_p$ norms that have the following form:
   \begin{equation}
     d_{L_p}(X,Y) = \Bigg( \sum_{i=1}^{t}(x_i - y_i)^p\Bigg)^{\frac{1}{p}},
   \end{equation}
   \noindent where $p$ is a positive integer \cite{yi00}. When $p=2$, we have the so-called Euclidean distance (ED). The $L_p$ norms have the advantage of being intuitive, parameter-free, and linear complexity to the length of the series for computing. The shortcoming is that these measures are sensitive to noise and misalignment in time because a fixed pairs of data points are compared. For this reason, these type of measures are called lock-step measures. In order to solve this problem, some elastic measures have been developed to allow time warping and, consequently, provide robust comparison results. Figure \ref{fig:align} illustrates a time series comparison made by lock-step and elastic measures, respectively. 

   \noindent The most famous elastic measures is the Dynamic Time Warping (DTW) that align two time series using the shortest warping path in a distance matrix \cite{berndt94}. A warping path $W$ defines a mapping consisting of a sequence of adjacent matrix. There is a high number of path combinations and the optimal path is the one that minimizes the global warping cost. The Short Time Series (STS) \cite{moller03} and DISSIM \cite{frentzos07} distances are designed to deal with time series collected in different sampling rates. The Complexity Invariant Distance (CID) \cite{batista14} calculates the Euclidean distance corrected by a complexity estimation of the series.

   \subsubsection{Edit Based distance measures} 

   Edit-based distances compute the distance between two series based on the minimum number of operations needed to transform one time series into another. This kind of measures is based on the string edit distance (levenshtein) that counts the number of character insertions, deletions and substitutions needed to transform one string into another. The Longest Common Subsequence (LCSS) \cite{vlachos02} is one of the best known edit based measures. It allows not only time warping, as DTW, but also gaps in comparison. Therefore, LCSS possesses two threshold parameters, $\varepsilon$ and $\delta$, for point matching and warping, respectively. 
   
   \subsubsection{Feature based distance measures} 

   This kind of measures has focus on extracting a number of features from the time series and comparing the extracted features instead of the raw data. Such features can be selected by various techniques, for example, using coefficients of a Wavelet Transform (DWT) as features \cite{zhang06}. In this category, the INTPER measure computes the distance based on the integrated periodogram from each series \cite{casado03} and, then, it uses the Pearson correlation (COR) \cite{golay98} to calculate the distance between time series. 

   \subsubsection{Structure based distance measures}

Different from feature based measures, structure based measures try to identify higher-level structures in the series. Some structure based measures use parametric models to represent the series, for example, Hidden Markov Models (HMM) \cite{smyth97} or ARMA \cite{xiong04}. In these cases, the similarity is measured by the probability of one modelled series produced by the underlying model of another. There are other measures, which use the concept of compression (CDM) \cite{keogh04}. The idea is that when concatenating and compressing two similar series, the compression ratio should be higher than the simple concatenation of them. 

\subsection{Time series clustering}
\label{subsec:ts_clustering_algs}

Clustering is one of the most common tasks in data mining. The goal is to divide data items into groups according a pre-defined similarity or distance measure. More specifically, clusters should maximize the intra-cluster similarity and minimize the inter-cluster similarity. In the context of time series data mining, the same idea applies. Considering a set of time series, the goal is to find groups of time series that are similar inside the cluster but are relatively different from times series of other clusters.

Time series clustering algorithms can be broadly classified into two approaches: data adaptation and algorithm adaptation \cite{liao05}. The former extracts features arrays from each time series data and, then, applies a conventional clustering algorithm. The latter modifies the traditional clustering algorithms in such a way that they can handle time series directly. Next, we review representative clustering methods following the above classification. 

\begin{itemize}

\item \textit{Time series clustering based on data adaptation}: This class of algorithms extracts some features of input time series and, then, apply traditional clustering algorithms without any change. The advantage of such an approach is that the feature extraction process can eventually reduce the amount of data and, consequently, reduce the processing time. Moreover, better results can be obtained if the characterization process is able to remove noise and filter out other kinds of irrelevant information. One shortcoming of this approach is the high number of parameters that the algorithms should handle. Guo et. al.\ \cite{guo08} present a technique that converts the raw data into a low dimensional array using independent component analysis and, then, apply $k$-means for clustering. Zakaria et al.\ \cite{zakaria12} propose an algorithm that firstly extracts sub-sequences called shapelets, which are local patterns in a time series and are highly predictive of a group. Then, the authors use the $k$-means algorithm to cluster shapelets. Brandmaier \cite{brandmaier11} introduces a method called Permutation Distribution Clustering (PDC) that makes an embedding of each time series into an $m$-dimensional space. The permutation distribution is obtained by counting the frequency of distinct order patterns in an $m$-embedding of the original time series. The embedding dimension $m$ is automatically chosen by PDC making it a parameter-free algorithm. The difference between time series is measured by the differences between their permutation distribution. After calculating this difference for each pair of time series, a hierarchical clustering algorithm, like single-linkage or complete-linkage, is applied to group similar series. 

\item \textit{Time series clustering based on algorithm adaptation}: This class of algorithms adapts traditional clustering algorithms to deal with time series. The major modification is the distance function that should be capable of distinguishing time series. For this purpose, various time series similarity measures can be used in distance-based clustering algorithms. The problem of this kind of algorithms is that the similarity measures usually consider all of the values, even outliers and noise, in the series. Since all the data points are involved in the similarity calculating, this approach demands much processing time and, thus, becomes infeasible to larger datasets. Golay et al.\ \cite{golay98} applied the fuzzy c-means algorithm to group time series extracted from functional MRI data. Maharaj \cite{maharaj00} proposed a method based on hypotheses testing. It considers that two time series are different if they have significantly different generating processes. Instead of building a distance matrix $D$, this method constructs a $P$ matrix where $p_{ij}$ corresponds to $p$-value obtained by testing if $X_i$ and $X_j$ were generated by the same model. The clustering algorithm groups together time series that have $p$-values greater than a significance level $\alpha$ previously specified by the user. Other adapted algorithms include Self-Organizing Maps (SOM) \cite{chappelier96}, Hidden Markov Models (HMM) \cite{smyth97} and Expectation Maximization (EM) \cite{yimin02}.

\end{itemize}

To our knowledge, there isn't work in the literature, which uses network community detection algorithms for time series clustering. The idea of using network theory to cluster time series was first presented by Zhang et al.\ \cite{zhang11}. The method consists of the construction of a network where each time series is represented by a vertex and each vertex is connected to its most similar one using DTW. Rather than clustering all vertices, this method selects some candidates (vertices with high degree) and considers that their neighbors belong to the same cluster. The authors proposed a hierarchical clustering that uses an DTW-based function that measures the similarity between clusters and iteratively merge the most similar ones. 

As having been mentioned in the Introduction section, network representation has definite advantage for characterizing global relationship among data samples and such an attractive feature is far from well explored in time series analysis. For this reason, we here conduct a comprehensive study on time series clustering using network representation. Specifically, we apply community detection algorithms to produce time series clusters. Computer simulations show that our approach has good performance. Moreover, it has the ability to identify arbitrary shape of clusters. 

\subsection{Community detection in networks}

Network (or graph) is one of the most powerful mechanisms to represent objects and their interactions or relations. Formally, a network is defined as follows.

\begin{definition}[Network]
    A network (or a graph) $G(V,E)$ is composed by a set of $n$ vertices $V=\{v_1,\dots,v_n\}$ and a set of $m$ edges $E=\{(v_i,v_j)\ | \ v_i,v_j\in V\}$ where $(v_i,v_j)$ is an edge that connects two vertices $v_i$ and $v_j$.
\end{definition}

Many real world systems are naturally represented as networks. Examples include social networks, protein interaction networks, neural networks and many others \cite{boccaletti06}. In data analysis domain, networks can be artificially constructed from the vector-based data format. One of the common ways to construct a network requires only a distance measure between the data samples in their original dataset. In this case, each sample is represented as a vertex and it is connected to its $k$ most similar ones. Such networks are called $k$-nearest neighbor networks ($k$-NN). In a similar way, network can be also constructed considering a threshold value $\varepsilon$. In this case, each pair of nodes is connected if the similarity between them is higher than $\varepsilon$. The networks constructed in this manner are called $\varepsilon$-nearest neighbor networks ($\varepsilon$-NN).

Communities are groups of highly connected vertices, while the connections between groups are sparse (Fig. \ref{fig:clustering_time_series}). Such structures are commonly observed in real world networks \cite{newman02}. Community detection is a task that involves searching for the cluster structure of vertices in a given network. It is not a trivial task, since evaluating all clustering (partitions) possibilities is NP-hard problem \cite{fortunato10}. Because of this difficulty, many algorithms have been proposed to find out reasonable network partitions in an efficient way. 

Several algorithms have been developed based on a network measure called modularity score, which measures how good is a particular partition of a network. In the Fast Greedy (FG) algorithm \cite{clauset04}, firstly, all edges are removed and each node itself is considered as a community. At each iteration, the algorithm determines which of the original edge, if it is added to this network, would generate the highest increase of the modularity. Then, this edge is inserted into the network and the two vertices (or communities) are merged. This process continues until all communities are merged resulting in just one community. Each iteration of the algorithm generates a possible solution but the best partition is that one with the highest modularity. The Multilevel (ML) algorithm \cite{blondel08} performs in the same way as FG, except that it does not stop when the highest modularity is found. After that, each community is abstracted to a single vertex and the process starts again with the merged communities. The process stops when there is just one vertex in the network.

Many other algorithms have been proposed using random walks to find communities. The idea is that short random walks in the network tend to stay in the same community. The Walktrap (WT) algorithm \cite{pons05} uses the same greedy strategy as FG and ML; however, it chooses the communities to be merged using a distance between vertices instead of using modularity. The distance is based on the probability distribution of a specific vertex reaches each of the other ones in a random walk of length $t$. If two vertices are in the same community, their probability distributions should be similar and their distance tends to be 0. The authors also make a generalization of the distance to communities and, at each iteration, the algorithm merges those two communities, which minimize the mean of the squared distances between each vertex and its community. 

Besides of the above mentioned algorithms, other strategies have also been considered to perform community detection. For example, the Label Propagation (LP) \cite{raghavan07} algorithm uses the concept of information diffusion in the network. It starts by giving a unique label to each vertex. At each iteration, all vertices are visited in a random sequence and each one receives the label with the highest occurrence of its neighbors. During the process, some labels disappear and others dominate. The algorithm converges when the label of each vertex of the network is the label of the majority of its neighbors. Finally, the communities are formed by vertices that share the same label. The Infomap (IM) algorithm \cite{rosvall08} use the concept of random walks and information diffusion. The idea is compressing the description of information flows in the network described by the trajectory of random walk. The result is a map that is a simplification of the network and highlight important structures (communities) of the network. For a full review on community detection algorithms, we refer the interested reader to \cite{fortunato10}.

% ==============================================================================
% Section -
% ==============================================================================
\section{Description of the proposed method}
\label{sec:method}

\begin{figure}[ht]
  \centering
  \includegraphics[width=.5\textwidth]{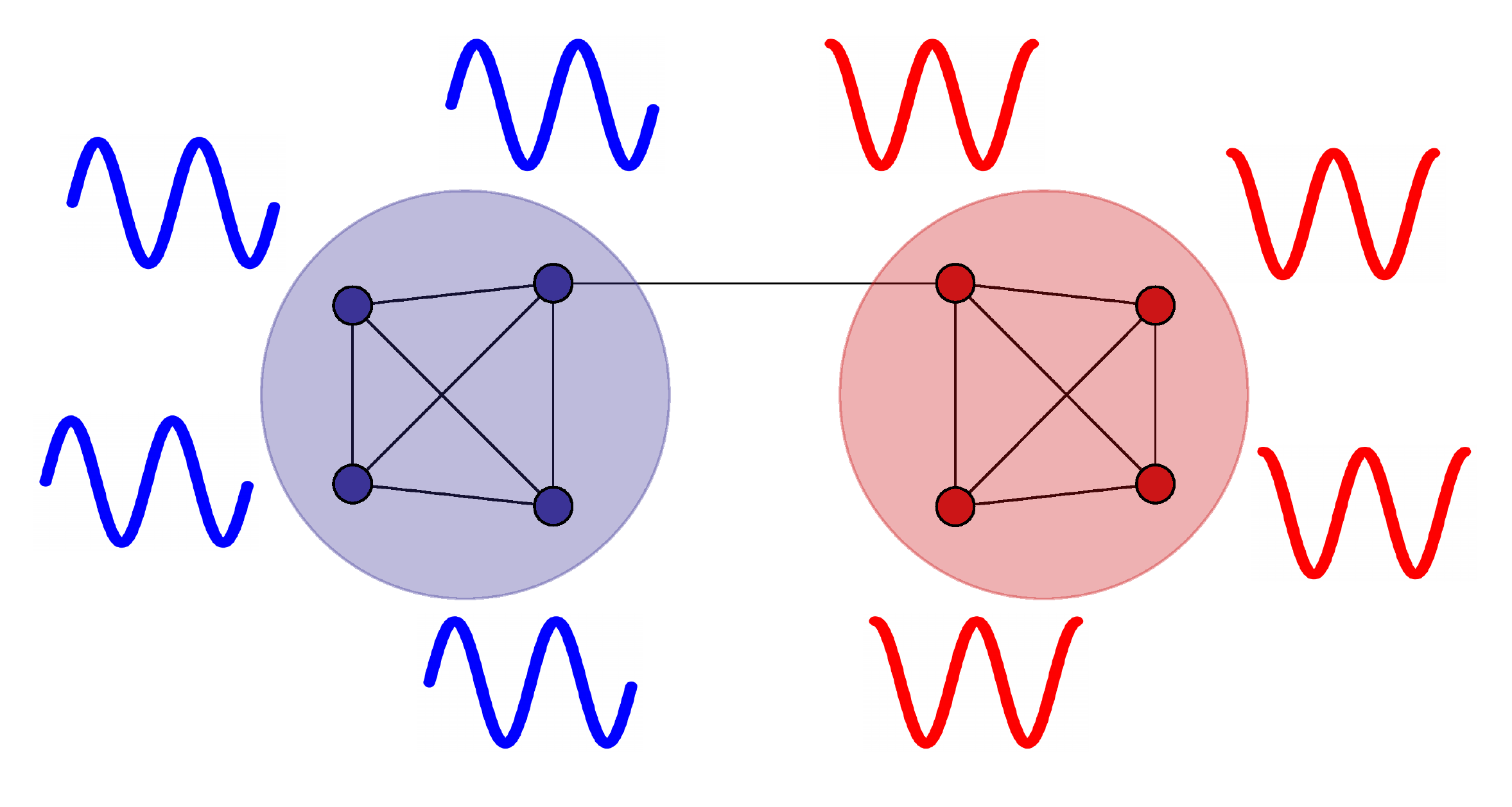}
  \caption{Time series clustering using community detection in networks. First, we construct a network where every vertex represents a time series connected to its most similar time series using a distance function. Then we apply community detection algorithms in order to cluster time series. The communities are represented by vertices with different colors.}
  \label{fig:clustering_time_series}
\end{figure}

The intuition behind our algorithm is simple. Each time series from a database is represented by a vertex and a distance measure is used to determine the similarity among time series and connect the most similar ones. As expected, similar time series tend to connect to each other and form communities. Thus, we can apply community detection algorithms to detect time series clusters. The idea of this algorithm is illustrated by Figure \ref{fig:clustering_time_series} and the whole process will be detailed in the following.

More specifically, the proposed method is performed in 4 steps: 1) data normalization, 2) time series distance calculation, 3) network construction and 4) community detection. Each step is described as follows:

\begin{enumerate}
  \item \textit{Normalization}: The first step is a pre-processing stage that intends to scale the dataset. As observed in \cite{keogh13}, normalization improves the search of similar time series when they have similar shapes but have different scales.

  \item \textit{Distance measures}: The second step consists of calculating the distance for each pair of time series in the data set and construct a distance matrix $D$, where $d_{ij}$ is the distance between series $X_i$ and $X_J$. A good choice of distance measure has strong influence on the network construction and clustering result.

  \item \textit{Network construction}: This steps intends to transform the matrix $D$ into a network.  In general, the two most used methods for network construction from a dataset are the $k$-NN and $\varepsilon$-NN. The way how the network is constructed highly affects the clustering result.

  \item \textit{Community detection}: After the network is constructed, we apply community detection algorithms in order to search for groups of densely connected vertices to form communities. There are plenty of community detection algorithms that use different strategies and the correct choosing again affects the clustering result.
\end{enumerate}

All these steps are presented in Algorithm \ref{alg:com_detection}.

\begin{algorithm}[!h]
% \SetLine
\caption{Time series clustering}
\label{alg:com_detection}
\SetKwInOut{Input}{input}
\Input{$dataset, k$ or $\varepsilon$}
\Begin{
  \texttt{normalization(}$dataset$\texttt{)}\;
  $D \gets$ \texttt{distanceMatrix(}$dataset$\texttt{)}\;
  $G \gets$ \texttt{netConstruction(}$D$, $k$ or $\varepsilon$ \texttt{)}\;
  $C \gets$ \texttt{communityDetection(}$G$\texttt{)}\;  
}
\end{algorithm}

The time complexity is defined as the sum of the complexities of each step of the method and it depends on the chosen algorithms and measures. Considering a dataset composed by $n$ time series all of length $t$, the z-score normalization of the dataset can be performed in O($nt$). Also considering that a time series measure can be calculated in a linear time (Table \ref{tab:measures}), the network construction needs O($n^2t$) computations. The time complexities for the community detection algorithms (Table \ref{tab:community}) are usually lower than quadratic and even can be linear \cite{SilvaZhao2012}; therefore, the complexity order of the proposed method is O($n^2t$).

Notice that the most time-consuming process is calculating the distances between all pais of data points, which is O($n^2t$). Therefore, any improvement of the nearest neighbor methods can be implemented in our method to reduce the computation time. For example, in Ref. \cite{chen09}, the authors proposes a divide and conquer method based on Lanczos Bisection for constructing a \emph{k}NN graph with complexity bounded by O($nt$). Using this improvement, the complexity order of the proposed time series clustering algorithm is reduced to O($nt$).  

% ==============================================================================
% Section -
% ==============================================================================
\section{Experimental evaluation}
\label{sec:experiments}

In this section, we present experimental results using the proposed methods. In order to make reproducibility easier, we provide a web page containing the source code of our algorithm \cite{extra15}. The experiments intend to find out the influence of the distance functions, network construction methods and community detection algorithms on time series data clustering. Finally, we compare our method to rival ones.

\subsection{Experiment settings}

For the experiments performed in this paper, we use 45 time series data sets from the UCR repository \cite{ucr14}. These data sets are described in \ref{append:data_set}. The experiments has objective to check the performance of each combination of time series distance measures (Tab. \ref{tab:measures}), networks construction methods ($\varepsilon$-NN or $k$-NN) and community detection algorithms (Tab. \ref{tab:community}) to each data sets. To compare the results, we use the Rand Index (RI) \cite{halkidi01} that measures the percentage of correct decisions made by the algorithms. The RI is defined as:

\begin{equation}
  RI = \frac{TP+TN}{n(n-1)/2}, 
\end{equation}

\noindent where $TP$ (true positive) is the numbers of pairs of time series that are correctly put in the same cluster, $TN$ (true negative) is the number of pairs that are correctly put in different clusters and $n$ is size of the data set. The RI for each clustering method is calculated comparing its result to the correct clustering (labels) provided by the UCR.

We will vary the parameters to find out the best clustering result, characterized by the RI index, for each data set. In the methods using $k$-NN, the best RI is achieved by varying parameter $k$ from 1 to $n-1$. In the methods using $\varepsilon$-NN, the best RI is achieved by varying $\varepsilon$ from $min(D)$ to $max(D)$ in 100 steps of $(max(D)-min(D))/100$, where $D$ is the distance matrix. For a fair comparison, the same procedure is considered in the rival methods. 

The results are presented using box plots that use rectangles to represent the middle half of the data divided by the median, represented by a black horizontal line. The vertical lines represent the max and min values. Black dots inside the boxes represent the mean values. Black dots outside the boxes represent the outlier values. For comparison purpose, we use non-parametric hypothesis tests according to \cite{demsar06} and provide the $p$-values for the reader interpretation. In all the cases, we consider a significance level of .05, i.e., $p$-values $\leq .05$ indicates a strong evidence that one method statistically better (or worse) than another. On the other hand, $p$-values close to 1 indicates that the algorithms under comparison are statistically equivalent. 

\begin{table}[ht]
  \centering
  % \footnotesize
  \begin{threeparttable}
  \caption{Time series distance functions used in the experiments} 
  \label{tab:measures}
      % \rowcolors{2}{gray!25}{white}
  \begin{tabular}{llc}
    \toprule
    Distance                                  & Cost        & Ref.           \\
    \midrule        
    Manhattan ($L_1$)                         & O($t$)      & \cite{yi00}         \\
    Euclidean (ED)                            & O($t$)      & \cite{yi00}         \\
    Infinite Norm ($L_\infty$)                & O($t$)      & \cite{yi00}         \\
    Dynamic Time Warp (DTW)                   & O($t^2$)    & \cite{berndt94}     \\
    Short Time Series (STS)                   & O($t$)      & \cite{moller03}     \\
    DISSIM                                    & O($t^2$)    & \cite{frentzos07}   \\
    Complexity-Invariant (CID)                & O($t$)      & \cite{batista14}    \\
    Wavelet Transform (DWT)                   & O($t$)      & \cite{zhang06}      \\
    Pearson Correlation (COR)                 & O($t$)       & \cite{golay98}      \\
    Integrated Periodogram (INTPER)           & O($t$)      & \cite{casado03}      \\
    \bottomrule
  \end{tabular}
    \begin{tablenotes}
      \footnotesize
      \item $t$ is the length of the series;
    \end{tablenotes}
  \end{threeparttable}
\end{table}

\begin{table}[ht]
  \centering
  % \footnotesize
  \begin{threeparttable}
    \caption{Community detection algorithms used in the experiments}
    \label{tab:community}
    % \rowcolors{2}{gray!25}{white}
    \begin{tabular}{llc}
    \toprule  
    Algorithm              & Cost               & Ref.       \\
    \midrule
    Fast Greedy (FG)       & O($n.\log^{2}{n}$) & \cite{clauset04} \\
    Multilevel (ML)        & O($m$)             & \cite{blondel08} \\
    Walktrap (WT)\tnote{a} & O($n^2.\log{n}$)   & \cite{pons05}    \\
    Infomap (IM)\tnote{b}  & O($m$)             & \cite{rosvall08} \\
    Label Propagation (LP) & O($m+n$)           & \cite{raghavan07} \\
    \bottomrule
    \end{tabular}
    \begin{tablenotes}
      \footnotesize
      \item[a] Walk length = 4;
      \item[b] Num. of trials to partition the network $nb.trials = 10$.
    \end{tablenotes}
  \end{threeparttable}
\end{table}

\subsection{Network construction influence}
\label{subsec:network_construction_influence}

\begin{figure}[h]
\centering
\subfloat[]{\label{fig:sim1_knn} \includegraphics[scale=0.3]{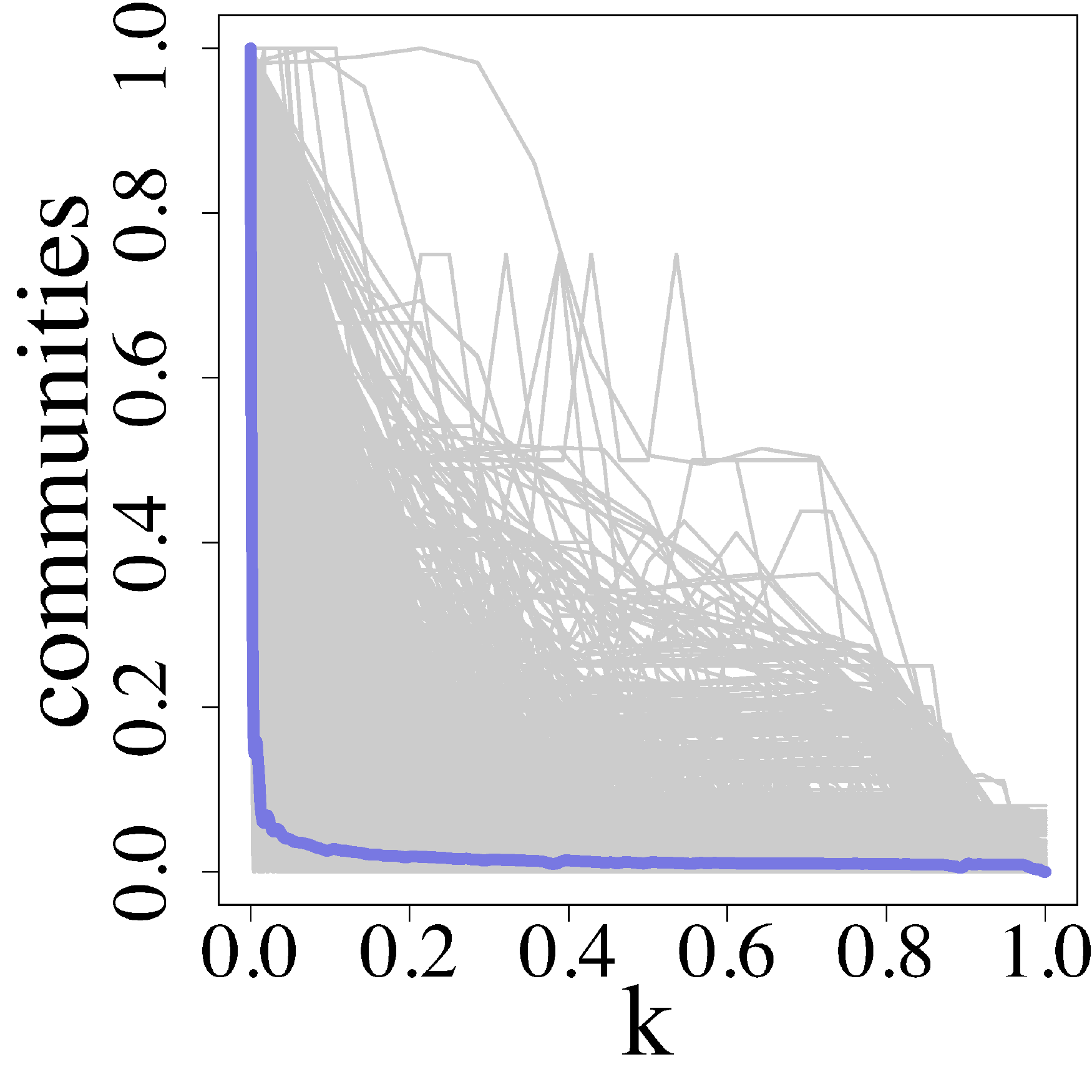}} 
\subfloat[]{\label{fig:sim1_eps} \includegraphics[scale=0.3]{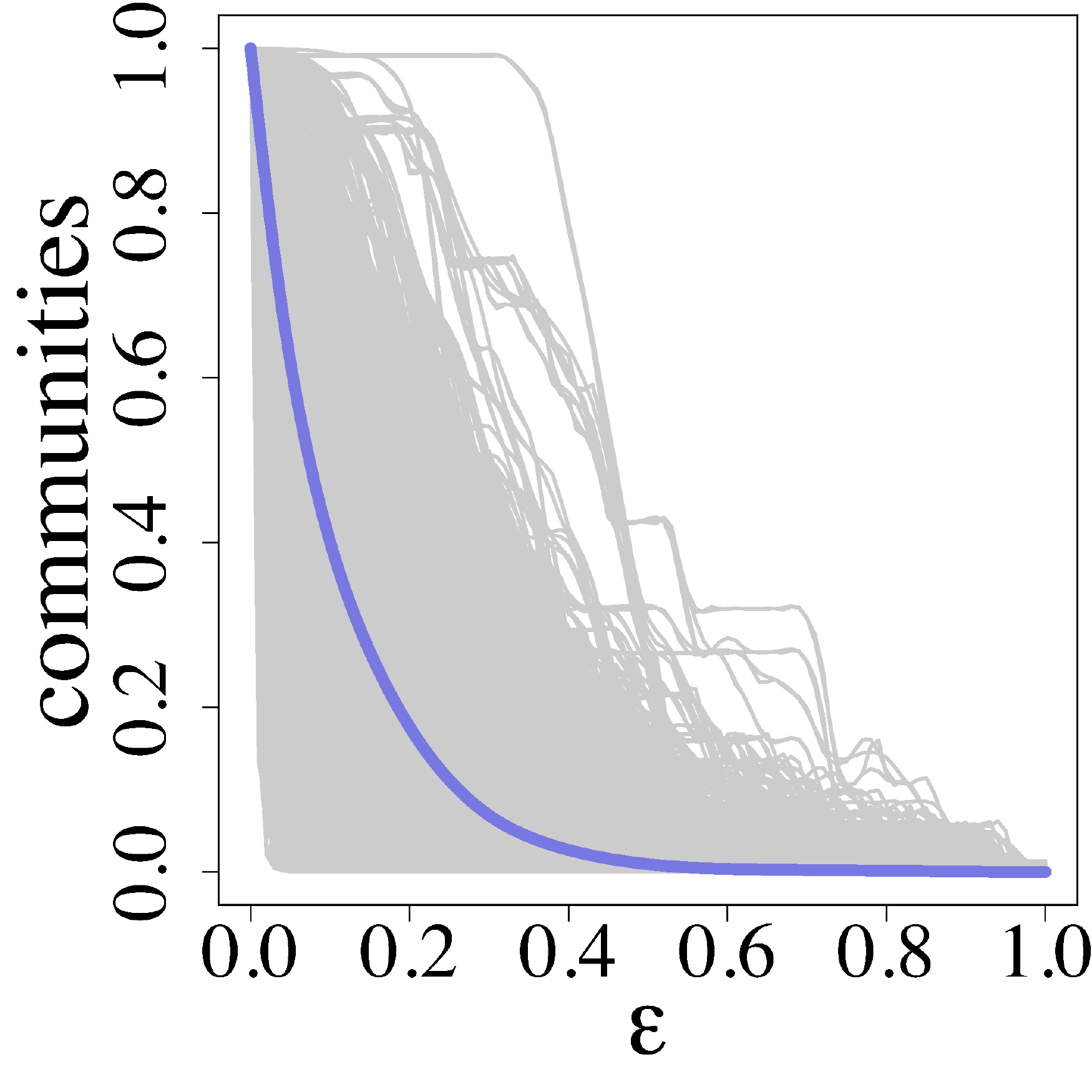}}
\caption{Influence of the parameters \protect\subref{fig:sim1_knn} $k$ and \protect\subref{fig:sim1_eps} $\varepsilon$ on the resulting number of communities. Weak (gray) lines represent the normalized real variation of the parameter for each combination of data sets, time series measures and community detection algorithms. The strongest line (blue) is a interpolation of all results, showing the average behavior. The $k$-NN construction method just allows discrete values of $k$ while the $\varepsilon$-NN method accepts continuous values. This difference explains why the $k$-NN interpolation presents the sharpest decrease. In small datasets, $k$ can assume just a few values and it makes that small variations of $k$ can result in a densely connected network.}
\label{fig:sim1} 
\end{figure}

The first experiment consists of evaluating the influence of the network construction on the community detection process. We verify how the parameters $k$ and $\varepsilon$ from the $k$-NN and $\varepsilon$-NN methods influence the network construction in order to provide a good strategy for correctly choosing these parameter and therefore get good clustering results. We start by running our method for all combinations of data sets, time series distance measures and community detection algorithms for various values of $k$ and $\varepsilon$. The results are shown in Figure \ref{fig:sim1}. 

\begin{figure*}[ht] %[!b]
\centering
\subfloat[]{\label{fig:knn_net1} \includegraphics[scale=0.2]{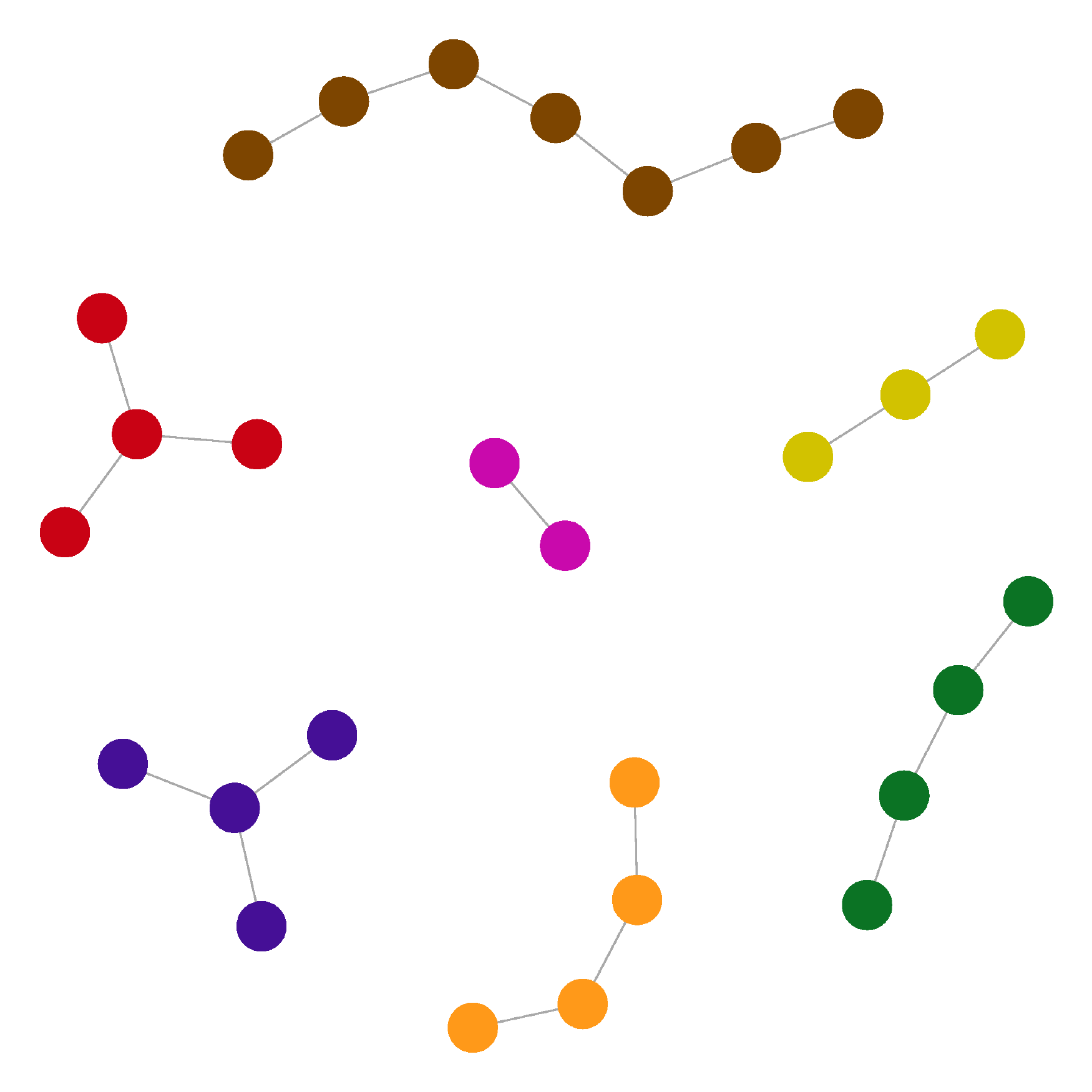}} \qquad
\subfloat[]{\label{fig:knn_net2} \includegraphics[scale=0.2]{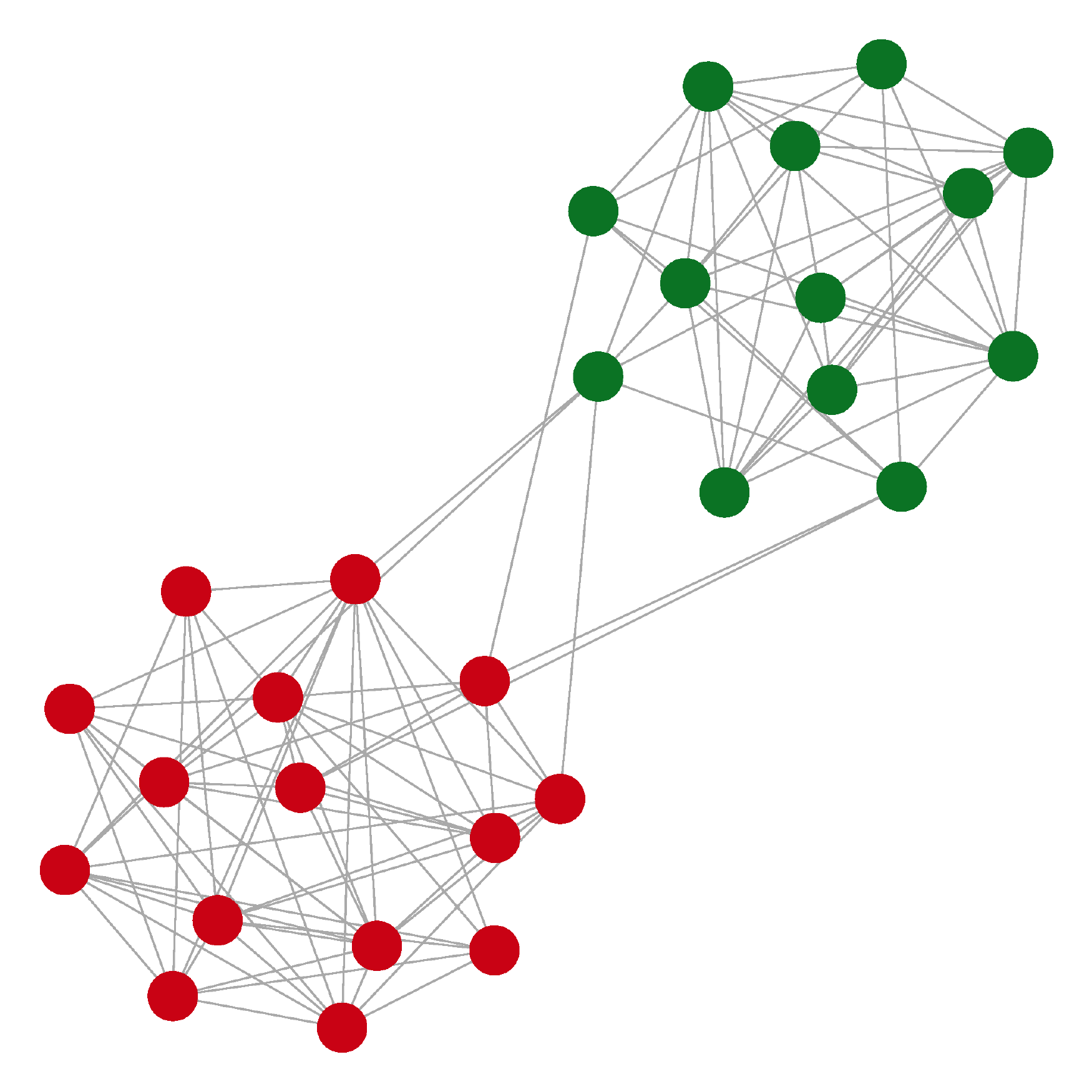}} \qquad
\subfloat[]{\label{fig:knn_net3} \includegraphics[scale=0.2]{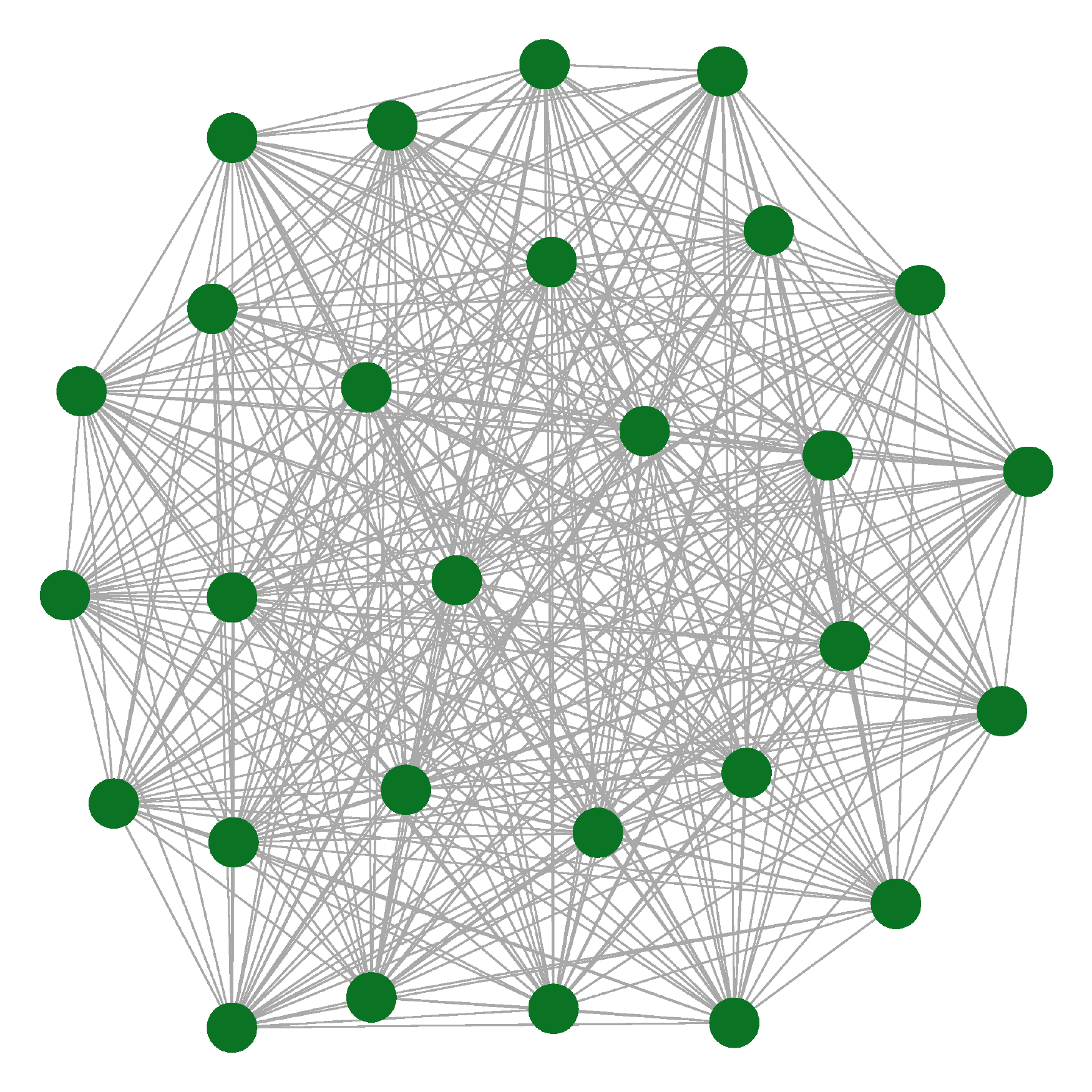}}
\caption{Example of the influence of the network construction method on the clustering result for the coffee data set (28 time series divided in 2 classes). In this case, we constructed three $k$-NN networks with the INTPER measure. Vertices colors represent the communities found with the fast greedy algorithm. \protect\subref{fig:knn_net1} $k=1$ results in a disconnected network where every component is a community (RI=0.64). \protect\subref{fig:knn_net2} $k=7$ creates a connected network with 2 communities that correctly clustered all the time series (RI=1). \protect\subref{fig:knn_net3} $k=27$ creates a fully connected network where the whole network form just one big community (RI=0.48).}
\label{fig:knn_net} 
\end{figure*}

When $k$ and $\varepsilon$ are small, vertices tend to make just few connections, which, in turn, generate many network components (a component is a connected subgraphs). As a result, community detection algorithms will produce a high number of clusters. On the other hand, if $k$ and $\varepsilon$ are high enough, all pairs of vertices tend to be connected, leading to a fully-connected network. In this case, all vertices are considered in one big community. Examples of these behaviors are depicted in Fig. \ref{fig:knn_net}. So the best clustering are usually achieved when intermediate values of $k$ and $\varepsilon$ are chosen. 

We also would like to check which method is better between $k$-NN and $\varepsilon$-NN. In the following experiment, we compare the best rand results achieved with both methods for each combination of datasets, distance measures and community detection algorithms. Tab. \ref{tab:best_rand_knn_eps} shows some statistics of the clustering results using the two different methods. Using the Wilcoxon signed-rank test (one-tailed) \cite{demsar06}, we conclude that, at a significance level of .05, the $\varepsilon$-NN method presents larger rand indexes ($p$-value $\leq .0001$), indicating that it is a better method.

\begin{table}[ht]
  \centering
  % \footnotesize
  % \scriptsize
  \begin{threeparttable}
    \caption{Performance of the two network construction methods.}
    \label{tab:best_rand_knn_eps}
    % \rowcolors{2}{gray!25}{white}
    \begin{tabular}{cccc}
    \toprule
      Network  & \multicolumn{3}{c}{Rand Index} \\ \cline{2-4}
      Method   &  Median  & Mean  & Std \\ 
    \midrule
         \textbf{$\varepsilon$-NN}  & \textbf{0.8436}  & \textbf{0.8133} & \textbf{0.1284} \\
         $k$-NN            & 0.8256  & 0.8012 & 0.1335 \\
    \bottomrule
    \end{tabular}
  \end{threeparttable}
\end{table}

\subsection{Time series distance function influence}
\label{subsec:distance_function_influence}

\begin{figure}[ht]
  \centering
  \includegraphics[width=.6\textwidth]{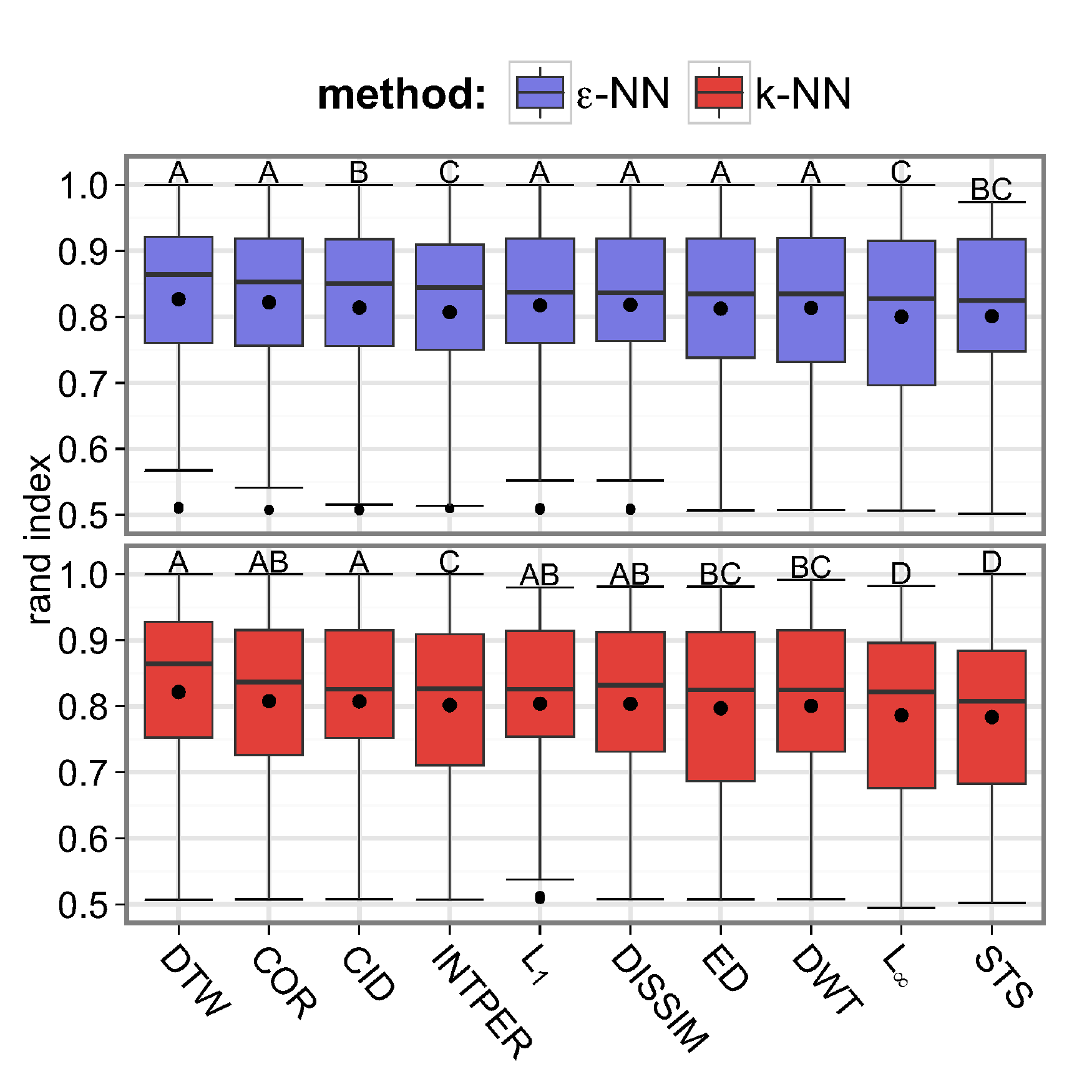}
  \caption{Box plot with the best rand distribution divided by measures and networks construction method. Measures with different letters (A, B, C or D) mean that they presented significant difference using the Nemenyi test and a significance level of .05.}
  \label{fig:boxplot_measures}
\end{figure}

Another factor, which may influence the performance of our method, is the time series distance function. Thus, we conduct studies to verify which one is the best for the clustering technique presented in this paper. For this purpose, we group the results by distance measures and plotted a boxplot. The results are shown in Figure \ref{fig:boxplot_measures}. 

According to the Friedman test, for both network construction methods, clustering results using different distance measures are significantly different ($p$-value $\leq .0001$). Hence, we proceed to the Nemenyi test to search for groups of similar measures. The real $p$-values are available in \cite{extra15}. According to the results, DTW measure presents the best results for both network construction methods. However, we cannot statistically affirm that it is a better measure. According to the Nemenyi test, we can affirm that, at a significance level of .05, $L_{\infty}$, STS and INTPER present worse results than other distance measures for both methods of network construction. 

\subsection{Community detection algorithm influence}
\label{subsec:community_detection_influence}

\begin{figure}[ht]
  \centering
  \includegraphics[width=.65\textwidth]{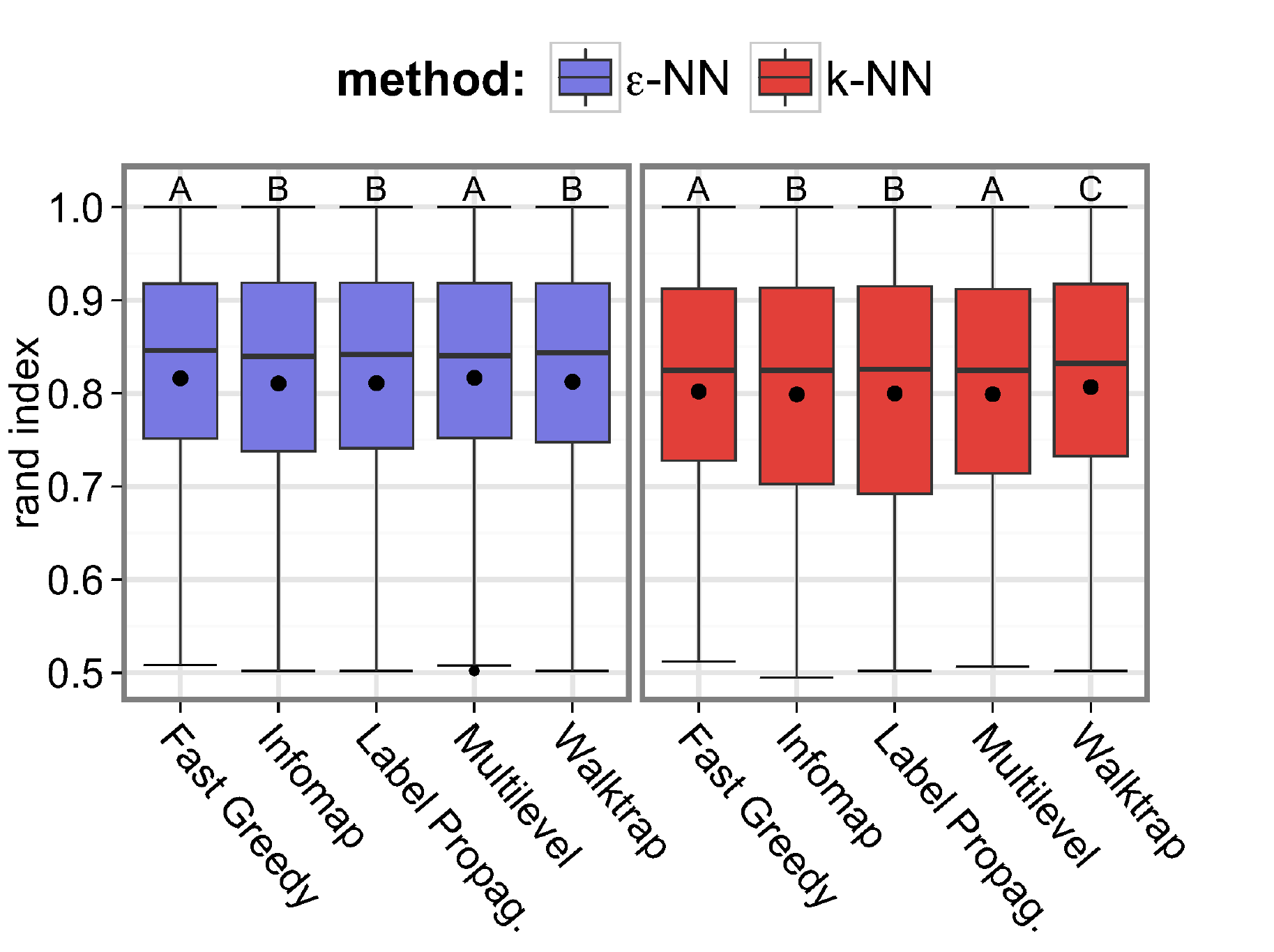}
  \caption{Box plot with the best rand distribution divided by community detection algorithms and networks construction method. Algorithms with different letters (A, B or C) mean that they presented significant difference using the Nemenyi test and a significance level of .05.}
  \label{fig:boxplot_commalgs}
\end{figure}

The third influence factor to our method is related to the community detection algorithm. Choosing a right algorithm can lead to better clustering results. So, we here verify which community detection algorithm is better for time series data clustering. For each combination of datasets and distance measures, we calculate the best rand index for each algorithm and plot a box plot, shown in Figure \ref{fig:boxplot_commalgs}. The results are divided into two parts regarding the two network construction methods ($k$ and $\varepsilon$) and, apparently, seems to be similar for both methods. 

To check whether the algorithms really have similar performance, we use the Friedman test \cite{demsar06} to compare the 5 algorithms and check whether there is a significant difference in the results. We conclude that, at a significance level of .05, for both network construction methods, the algorithms do not present similar results ($p$-value $\leq .0001$). Thus, the next step of our analysis consists of making a post-hoc analysis to check the difference between the algorithms. In this case, we use the Nemenyi test \cite{demsar06} to compare pairs of algorithms. The real $p$-values are available in \cite{extra15}. For the $k$-NN method, we find  that the Walktrap algorithm is, at a significance level of .05, better than the others. For the $\varepsilon$-NN method, the results show that the Fast Greedy and multilevel algorithms present statistically similar results and these are better than the Infomap, label propagation and Walktrap algorithm.

\subsection{Comparison to rival Methods}
\label{subsec:comparison_rival_methods}

Now we present a comparison of our approach to other time series clustering methods. For this comparison, we chose the combination of network construction method, the distance function and the community detection algorithm , which leads to the best experimental results so far. The first step consists of evaluating which algorithm achieves the best median value. We opt to compare the median instead of average because it is less sensitive to outliers \cite{demsar06}. The result is presented in Tab. \ref{tab:best_rand_community}.

\begin{table}[ht]
  \centering
  % \footnotesize
  % \scriptsize
  \begin{threeparttable}
    \caption{Performance of different combinations of networks construction methods, distance functions and community detection algorithms.}
    \label{tab:best_rand_community}
    % \rowcolors{2}{gray!25}{white}
    \begin{tabular}{cccccc}
    \toprule
    Network        & Dist   & Community     & \multicolumn{3}{c}{Rand Index} \\ \cline{4-6}
    Method         & Func.  & Detect. Alg.  &  Median  & Mean  & Std \\ 
    \midrule
        \textbf{$\varepsilon$-NN} & \textbf{DTW}  & \textbf{multilevel} & \textbf{0.8671} & \textbf{0.8309} & \textbf{0.1309}\\
        $k$-NN           & DTW        & fastgreedy  & 0.8644            & 0.8207 & 0.1381 \\
        $k$-NN           & DTW        & walktrap    & 0.8644            & 0.8283 & 0.1297 \\
        $k$-NN           & DTW        & infomap     & 0.8642            & 0.8191 & 0.1431 \\
        $\varepsilon$-NN & DTW        & infomap     & 0.8642            & 0.8225 & 0.1360 \\
        $\varepsilon$-NN & DTW        & walktrap    & 0.8642            & 0.8281 & 0.1283 \\
        $\vdots$         & $\vdots$   & $\vdots$    & $\vdots$          & $\vdots$ & $\vdots$\\
        $k$-NN           & $L_\infty$ & label prop. & 0.8163            & 0.7831 & 0.1408 \\
        $k$-NN           & ED         & fastgreedy  & 0.8127            & 0.7958 & 0.1310 \\
        $k$-NN           & STS        & infomap     & 0.8073            & 0.7812 & 0.1415 \\
        $k$-NN           & STS        & fastgreedy  & 0.8016            & 0.7822 & 0.1294 \\
        $k$-NN           & STS        & multilevel  & 0.8016            & 0.7802 & 0.1307 \\
        $k$-NN           & STS        & label prop. & 0.7980            & 0.7751 & 0.1388 \\
    \bottomrule
    \end{tabular}
    \begin{tablenotes}
      \footnotesize
      % \centering
      \item Results were sorted by the median values
    \end{tablenotes}
  \end{threeparttable}
\end{table}

According to Tab. \ref{tab:best_rand_community}, the best results for the community detection approach is achieved by using the multilevel algorithm with the $\varepsilon$-NN construction method and the DTW distance function. This result confirms to all the studies of influences previously presented in this paper. 

For comparison purpose, we firstly consider some classic clustering algorithms: $k$-medoids, complete-linkage, single-linkage, average-linkage, median-linkage, centroid-linkage and diana \cite{gan07}. For a fair comparison, we firstly find out which distance function leads to the better results for each rival method. Once again, we use the median to rank the results, that are presented in Tab. \ref{tab:rand_clustering}.

\begin{table}[ht]
  \centering
  % \footnotesize
  % \scriptsize
  \begin{threeparttable}
    \caption{Best time series measures for each of the rival methods}
    \label{tab:rand_clustering}
    % \rowcolors{2}{gray!25}{white}
    \begin{tabular}{ccccc}
    \toprule
    Clustering           & Dist     & \multicolumn{3}{c}{Rand index} \\ \cline{3-5}
    Algorithm            & Func.    & Median  & Mean  & Std \\ 
    \midrule         
      Diana              & DTW      & 0.8596 & 0.8167 & 0.1369 \\
      Centroid Linkage   & DTW      & 0.8593 & 0.8075 & 0.1306 \\
      Single Linkage     & DTW      & 0.8593 & 0.8164 & 0.1320 \\
      Median Linkage     & DTW      & 0.8591 & 0.8075 & 0.1294 \\
      Average Linkage    & CID      & 0.8575 & 0.8138 & 0.1375 \\
      $k$-medoids (PAM)  & COR      & 0.8534 & 0.8113 & 0.1310 \\
      Complete Linkage   & DTW      & 0.8501 & 0.8214 & 0.1249 \\
    \bottomrule
    \end{tabular}
  \end{threeparttable}
\end{table}

\begin{figure}[ht]
  \centering
  \includegraphics[width=.6\textwidth]{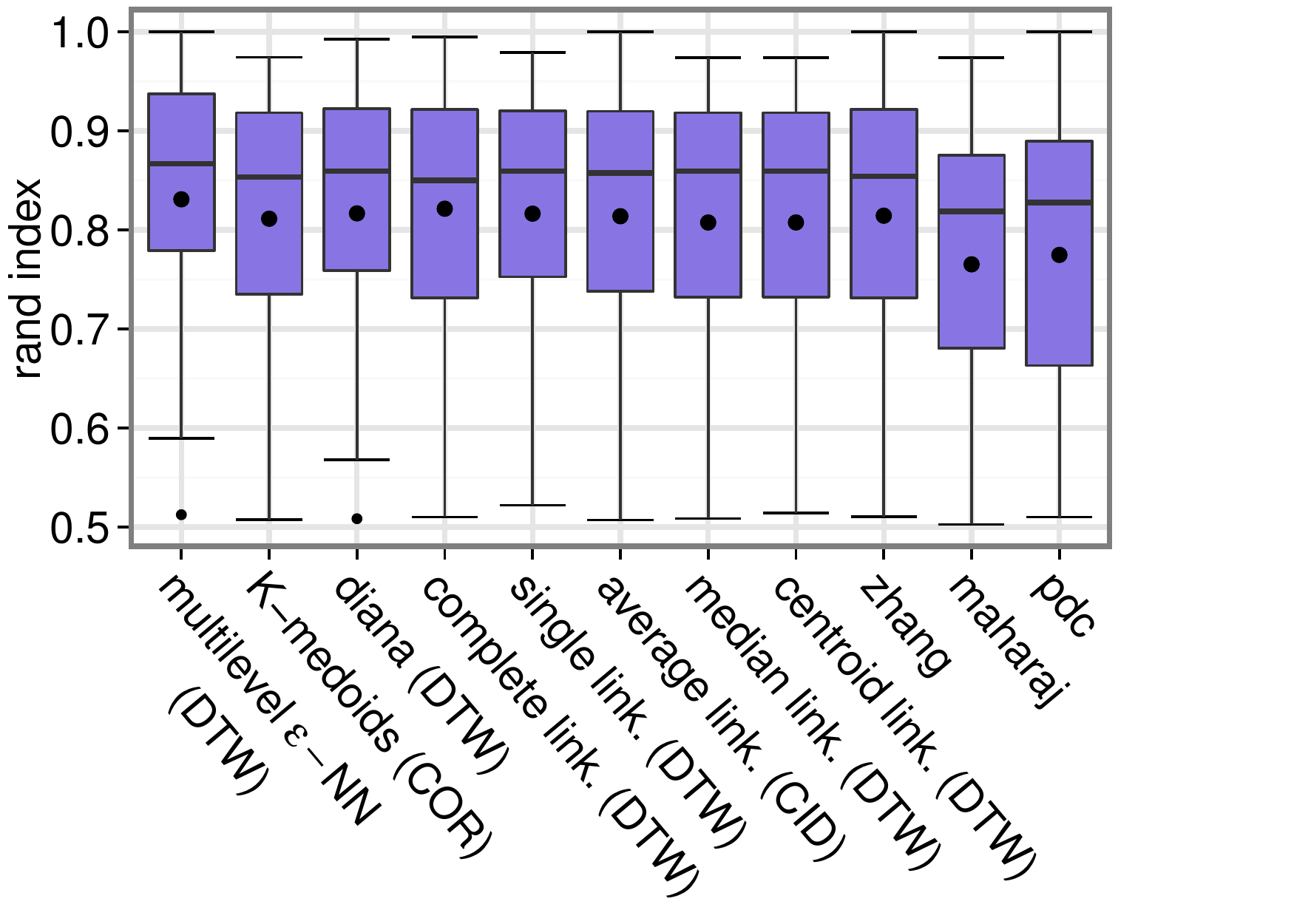}
  \caption{Box plot with the comparison of different time series clustering algorithms.}
  \label{fig:rand_clustering}
\end{figure}

Besides of those classic clustering algorithms, we also consider three up-to-date ones: Zhang's method \cite{zhang11}, Maharaj's method \cite{maharaj00} and PDC \cite{brandmaier11} (briefly described in Sec. \ref{subsec:ts_clustering_algs}). For Zhang's method, we vary the number of clustering candidates from 1 to the size of each dataset and report the best RI. In Maharaj's method, we search for the best RI varying the significance level $\alpha$ from 0 to 1 in steps of 0.5. For PDC, we use the complete linkage clustering algorithm and report the best RI from the hierarchy. Tables \ref{tab:rand_clustering_comparison1} and \ref{tab:rand_clustering_comparison2} show the best rand index for each algorithm and the corresponding data set. Figure \ref{fig:rand_clustering} summarizes this information in a box plot. 

\begin{table}[!h]
\centering
\scriptsize
\begin{threeparttable}
  \caption{Best rand index for each clustering algorithm I}
  \label{tab:rand_clustering_comparison1}
\begin{tabular}{rcccccc}
  \toprule
    & Multilevel        & $k$-Medoids & Diana & Complete & Single  & Average \\
    & $\varepsilon$-NN  & (COR)       & (DTW) & Linkage  & Linkage & Linkage  \\
    & (DTW)             &             &       & (DTW)    & (DTW)   & (CID)    \\
  \midrule 
  adiac & 0.97 & 0.97 & 0.97 & 0.97 & 0.97 & 0.97 \\
  beef & 0.83 & 0.83 & 0.83 & 0.83 & 0.83 & 0.83 \\
  car & 0.78 & 0.77 & 0.77 & 0.77 & 0.76 & 0.77 \\
  cbf & 0.96 & 0.73 & 0.88 & 0.92 & 0.82 & 0.85 \\
  chlorine\_concentration & 0.59 & 0.59 & 0.59 & 0.59 & 0.59 & 0.59 \\
  cinc\_ecg\_torso & 0.75 & 0.83 & 0.76 & 0.76 & 0.75 & 0.82 \\
  coffee & 0.60 & 0.86 & 0.58 & 0.58 & 0.62 & 0.58 \\
  cricket\_x & 0.92 & 0.92 & 0.92 & 0.92 & 0.92 & 0.92 \\
  cricket\_y & 0.92 & 0.92 & 0.92 & 0.92 & 0.92 & 0.92 \\
  cricket\_z & 0.92 & 0.92 & 0.92 & 0.92 & 0.92 & 0.92 \\
  diatom\_size\_reduction & 0.97 & 0.96 & 0.93 & 0.85 & 0.93 & 1.00 \\
  ecg\_five\_days & 0.63 & 0.55 & 0.63 & 0.68 & 0.60 & 0.55 \\
  ecg & 0.61 & 0.66 & 0.57 & 0.68 & 0.57 & 0.70 \\
  face\_all & 0.96 & 0.94 & 0.95 & 0.95 & 0.94 & 0.94 \\
  face\_four & 0.90 & 0.78 & 0.83 & 0.91 & 0.79 & 0.79 \\
  faces\_ucr & 0.94 & 0.92 & 0.94 & 0.95 & 0.92 & 0.92 \\
  fish & 0.87 & 0.87 & 0.87 & 0.87 & 0.87 & 0.88 \\
  gun & 0.60 & 0.60 & 0.57 & 0.59 & 0.60 & 0.63 \\
  haptics & 0.80 & 0.80 & 0.80 & 0.80 & 0.80 & 0.80 \\
  inlineskate & 0.86 & 0.86 & 0.86 & 0.86 & 0.86 & 0.86 \\
  italy\_power\_demand & 0.71 & 0.88 & 0.69 & 0.70 & 0.68 & 0.71 \\
  lighting2 & 0.64 & 0.55 & 0.57 & 0.70 & 0.61 & 0.55 \\
  lighting7 & 0.85 & 0.85 & 0.85 & 0.85 & 0.85 & 0.86 \\
  mallat & 0.94 & 0.95 & 0.97 & 0.97 & 0.91 & 0.94 \\
  medical\_images & 0.69 & 0.69 & 0.69 & 0.69 & 0.69 & 0.69 \\
  mote\_strain & 0.78 & 0.66 & 0.65 & 0.64 & 0.61 & 0.58 \\
  oliveoil & 0.88 & 0.91 & 0.86 & 0.82 & 0.87 & 0.90 \\
  osuleaf & 0.83 & 0.82 & 0.83 & 0.83 & 0.82 & 0.83 \\
  plane & 1.00 & 0.97 & 0.99 & 0.99 & 0.97 & 0.98 \\
  sony\_AIBO\_Robot\_surface\_ii & 0.83 & 0.74 & 0.77 & 0.74 & 0.82 & 0.74 \\
  sony\_AIBO\_Robot\_surface & 0.85 & 0.69 & 0.86 & 0.73 & 0.92 & 0.94 \\
  starlightcurves & 0.83 & 0.82 & 0.83 & 0.83 & 0.83 & 0.83 \\
  swedishleaf & 0.94 & 0.94 & 0.94 & 0.94 & 0.94 & 0.94 \\
  symbols & 0.97 & 0.95 & 0.97 & 0.97 & 0.97 & 0.96 \\
  synthetic\_control & 0.95 & 0.89 & 0.94 & 0.92 & 0.87 & 0.89 \\
  trace & 0.87 & 0.76 & 0.86 & 0.86 & 0.91 & 0.86 \\
  two\_lead\_ecg & 0.61 & 0.56 & 0.59 & 0.68 & 0.62 & 0.57 \\
  two\_patterns & 1.00 & 0.75 & 0.94 & 0.94 & 0.98 & 0.75 \\
  uwavegesturelibrary\_x & 0.89 & 0.88 & 0.88 & 0.88 & 0.89 & 0.89 \\
  uwavegesturelibrary\_y & 0.88 & 0.88 & 0.88 & 0.88 & 0.88 & 0.88 \\
  uwavegesturelibrary\_z & 0.88 & 0.88 & 0.88 & 0.88 & 0.89 & 0.89 \\
  wafer & 0.82 & 0.82 & 0.82 & 0.82 & 0.82 & 0.82 \\
  word\_synonyms & 0.91 & 0.91 & 0.92 & 0.91 & 0.92 & 0.92 \\
  words50 & 0.96 & 0.96 & 0.96 & 0.96 & 0.96 & 0.96 \\
  yoga & 0.51 & 0.51 & 0.51 & 0.51 & 0.52 & 0.51 \\
  \midrule 
  \textbf{Median} & 0.87 & 0.85 & 0.86 & 0.85 & 0.86 & 0.86 \\
  \textbf{Mean}   & 0.83 & 0.81 & 0.82 & 0.82 & 0.82 & 0.81 \\
  \textbf{St.D.}  & 0.13 & 0.13 & 0.14 & 0.12 & 0.13 & 0.14 \\
   \bottomrule
\end{tabular}
\end{threeparttable}
\end{table}

\begin{table}[!h]
\centering
\scriptsize
\begin{threeparttable}
  \caption{Best rand index for each clustering algorithm II}
  \label{tab:rand_clustering_comparison2}
\begin{tabular}{rccccc}
  \toprule
  & Median  & Centroid & Zhang          & Maharaj          & PDC \\
  & Linkage & Linkage  & \cite{zhang11} & \cite{maharaj00} & Comp. Link. \\
  & (DTW)   & (DTW)    &                &                  & \cite{brandmaier11} \\
  \midrule 
  adiac & 0.97 & 0.97 & 0.97 & 0.97 & 0.97 \\
  beef & 0.83 & 0.83 & 0.80 & 0.83 & 0.83 \\
  car & 0.77 & 0.78 & 0.76 & 0.76 & 0.76 \\
  cbf & 0.74 & 0.74 & 0.92 & 0.68 & 0.69 \\
  chlorine\_concentration & 0.59 & 0.59 & 0.59 & 0.59 & 0.59 \\
  cinc\_ecg\_torso & 0.75 & 0.75 & 0.75 & 0.75 & 0.89 \\
  coffee & 0.59 & 0.61 & 0.58 & 0.52 & 0.52 \\
  cricket\_x & 0.92 & 0.92 & 0.92 & 0.92 & 0.92 \\
  cricket\_y & 0.92 & 0.92 & 0.92 & 0.92 & 0.92 \\
  cricket\_z & 0.92 & 0.92 & 0.92 & 0.92 & 0.92 \\
  diatom\_size\_reduction & 0.87 & 0.87 & 0.95 & 0.74 & 0.76 \\
  ecg\_five\_days & 0.64 & 0.63 & 0.60 & 0.59 & 0.57 \\
  ecg & 0.57 & 0.60 & 0.56 & 0.68 & 0.57 \\
  face\_all & 0.93 & 0.93 & 0.95 & 0.93 & 0.93 \\
  face\_four & 0.79 & 0.78 & 0.82 & 0.75 & 0.76 \\
  faces\_ucr & 0.93 & 0.93 & 0.94 & 0.91 & 0.91 \\
  fish & 0.87 & 0.87 & 0.87 & 0.86 & 0.86 \\
  gun & 0.62 & 0.58 & 0.59 & 0.51 & 0.54 \\
  haptics & 0.80 & 0.80 & 0.79 & 0.80 & 0.80 \\
  inlineskate & 0.86 & 0.86 & 0.85 & 0.86 & 0.86 \\
  italy\_power\_demand & 0.64 & 0.62 & 0.70 & 0.53 & 0.52 \\
  lighting2 & 0.65 & 0.62 & 0.62 & 0.55 & 0.55 \\
  lighting7 & 0.85 & 0.85 & 0.86 & 0.84 & 0.84 \\
  mallat & 0.92 & 0.91 & 0.94 & 0.87 & 0.89 \\
  medical\_images & 0.69 & 0.69 & 0.69 & 0.69 & 0.68 \\
  mote\_strain & 0.63 & 0.63 & 0.62 & 0.53 & 0.66 \\
  oliveoil & 0.87 & 0.88 & 0.83 & 0.73 & 0.72 \\
  osuleaf & 0.82 & 0.83 & 0.82 & 0.82 & 0.82 \\
  plane & 0.95 & 0.95 & 1.00 & 0.86 & 0.86 \\
  sony\_AIBO\_Robot\_surface\_ii & 0.73 & 0.73 & 0.79 & 0.71 & 0.51 \\
  sony\_AIBO\_Robot\_surface & 0.92 & 0.92 & 0.73 & 0.72 & 0.86 \\
  starlightcurves & 0.81 & 0.83 & 0.81 & 0.57 & 0.62 \\
  swedishleaf & 0.94 & 0.94 & 0.94 & 0.93 & 0.94 \\
  symbols & 0.97 & 0.97 & 0.99 & 0.83 & 0.94 \\
  synthetic\_control & 0.86 & 0.86 & 0.94 & 0.84 & 0.84 \\
  trace & 0.86 & 0.87 & 0.87 & 0.84 & 0.75 \\
  two\_lead\_ecg & 0.58 & 0.58 & 0.58 & 0.52 & 0.55 \\
  two\_patterns & 0.90 & 0.92 & 0.98 & 0.75 & 0.75 \\
  uwavegesturelibrary\_x & 0.88 & 0.89 & 0.89 & 0.88 & 0.88 \\
  uwavegesturelibrary\_y & 0.88 & 0.88 & 0.88 & 0.88 & 0.88 \\
  uwavegesturelibrary\_z & 0.88 & 0.88 & 0.88 & 0.88 & 0.88 \\
  wafer & 0.83 & 0.82 & 0.82 & 0.82 & 1.00 \\
  word\_synonyms & 0.92 & 0.92 & 0.91 & 0.91 & 0.91 \\
  words50 & 0.96 & 0.97 & 0.96 & 0.96 & 0.96 \\
  yoga & 0.51 & 0.51 & 0.51 & 0.50 & 0.52 \\
  \midrule 
  \textbf{Median} & 0.86 & 0.86 & 0.85 & 0.82 & 0.83 \\
  \textbf{Mean}   & 0.81 & 0.81 & 0.81 & 0.77 & 0.77 \\
  \textbf{St.D.}  & 0.13 & 0.13 & 0.14 & 0.14 & 0.15 \\
   \bottomrule
\end{tabular}
\end{threeparttable}
\end{table}

We use the Wilcoxon paired test to compare our method to all other ones. To compensate the multiple pairwise comparison, we use the Holm-Bonferroni adjusting method \cite{demsar06}. At a significance level of .05, we conclude that the community detection approach presents better results ($p$-values $\leq .02$) than $k$-medoids (PAM), diana, median-linkage, centroid-linkage, Zhang's method \cite{zhang11}, Maharaj's method \cite{maharaj00} and PDC \cite{brandmaier11}. Even though our approach has presented higher median and mean values, we cannot conclude that it is statistically better than complete-linkage, single-linkage and average-linkage ($p$-values $\leq .32$) yet.

\subsection{Detecting time series clusters with time-shifts}
\label{subsec:efficiency_time_similarity}

\begin{figure}[ht] %[!b]
\centering
\subfloat[]{\label{fig:cbf_ds_1} \includegraphics[scale=0.3]{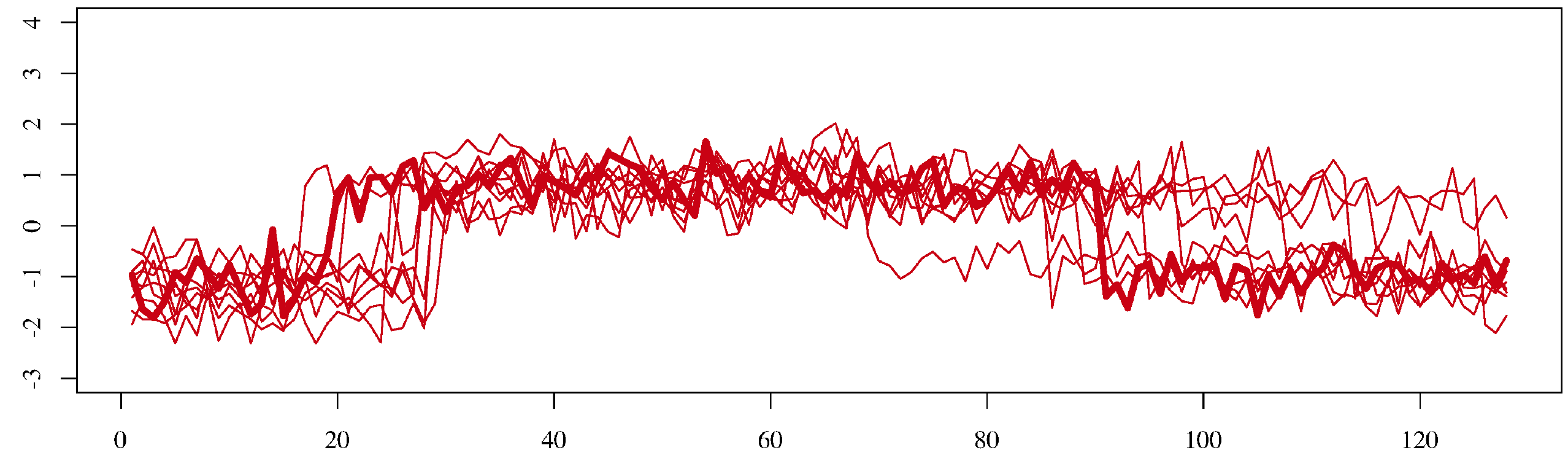}}\\ 
\subfloat[]{\label{fig:cbf_ds_2} \includegraphics[scale=0.3]{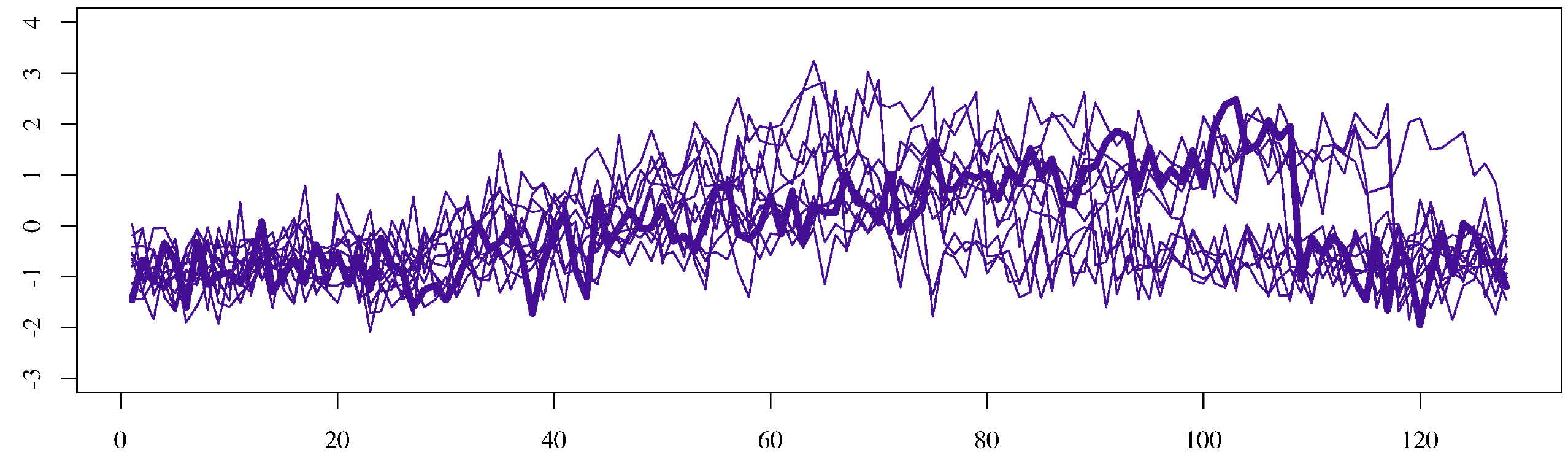}}\\
\subfloat[]{\label{fig:cbf_ds_3} \includegraphics[scale=0.3]{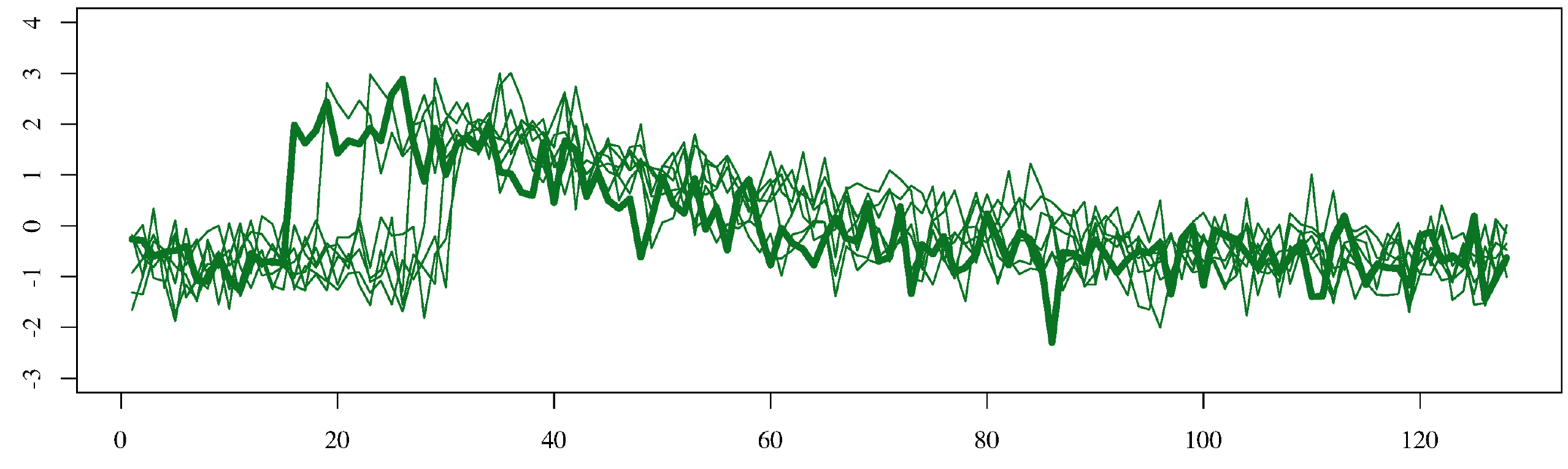}}
\caption{The Cylinder-Bell-Funnel (CBF) data set is composed by three groups of series: \protect\subref{fig:cbf_ds_1} the cylinder group of series is characterized by a plateau, \protect\subref{fig:cbf_ds_2} the bell group by an increasing linear ramp followed by a sharp decrease and \protect\subref{fig:cbf_ds_3} the funnel group by a sharp increase followed by a decreasing ramp.}
\label{fig:cbf_ds} 
\end{figure}

Clustering algorithms should be capable of detecting groups of time series that have similar variations in time. To exemplify the efficiency of our method in detecting similarity with time shifts, we consider the Cylinder-Bell-Funnel (CBF) data set, that is formed by 30 time series of length 128 divided into 3 groups \cite{geurts02}. Each group is defined by a specific pattern. The cylinder group of series is characterized by a plateau, the bell group by an increasing linear ramp followed by a sharp decrease and the funnel group by a sharp increase followed by a decreasing ramp. Even composed by a small number of time series, this data set presents characteristics that make difficult the detection of similarity. In this data set, the starting time, the duration and the amplitude patterns among the time series of the same group are different. A random Gaussian noise is also added to the series to reproduce the natural behavior. Figure \ref{fig:cbf_ds} shows the CBF data set.

\begin{figure}[!b]
  \centering
  \includegraphics[width=.45\textwidth]{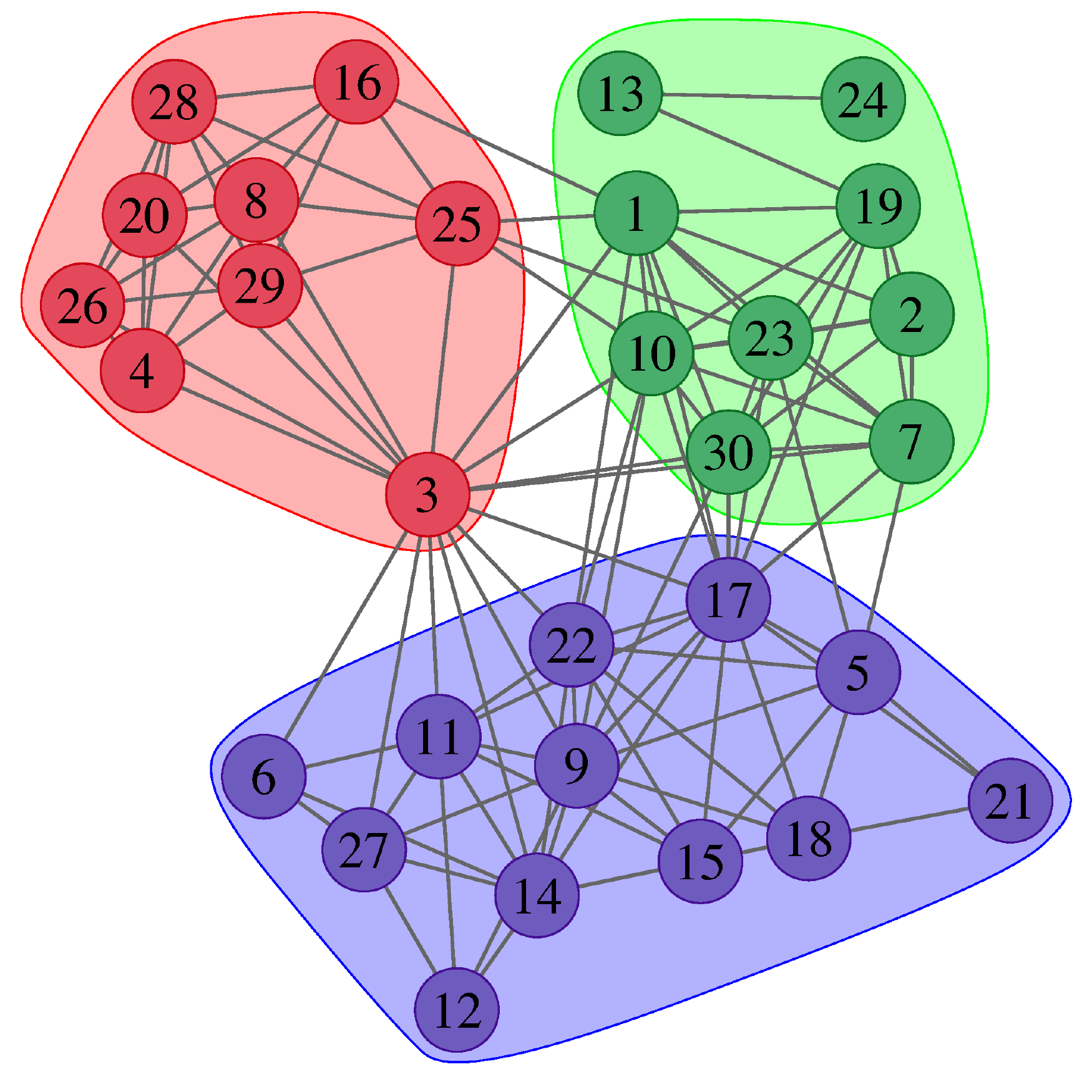}
  \caption{Network representation of the CBF data set using the $\varepsilon$-NN construction method ($\varepsilon=58.87$) with DTW. Vertices colors indicate the 3 communities that represent each group of time series illustrated in Fig. \ref{fig:cbf_ds}. All the 30 time series were correctly clustered, except one ($RI=0.96$). The time series with label ``3'' belongs to community of color blue (bottom).} 
  \label{fig:cbf_net}
\end{figure}

Using our approach, we build a $\varepsilon$-NN ($\varepsilon=58.87$) with DTW and then apply the multilevel community detection algorithm. The result (Fig. \ref{fig:cbf_net}) is a network with 3 communities, each one representing an original cluster of the data set. Our approach correctly finds out all the time series clusters, except the one with label ``3'' in Fig. \ref{fig:cbf_net}. In this simulation, we get $RI=0.96$ for our method. The rival method (Tab. \ref{tab:rand_clustering}) that achieves the best clustering result for this data set is the complete linkage with DTW: $RI=0.91$.

\subsection{Efficiency to detect shape patterns}
\label{subsec:efficiency_shape_patterns}

\begin{figure}[t] %[!b]
\centering
\subfloat[]{\label{fig:two_patterns_ds_1} \includegraphics[scale=0.16]{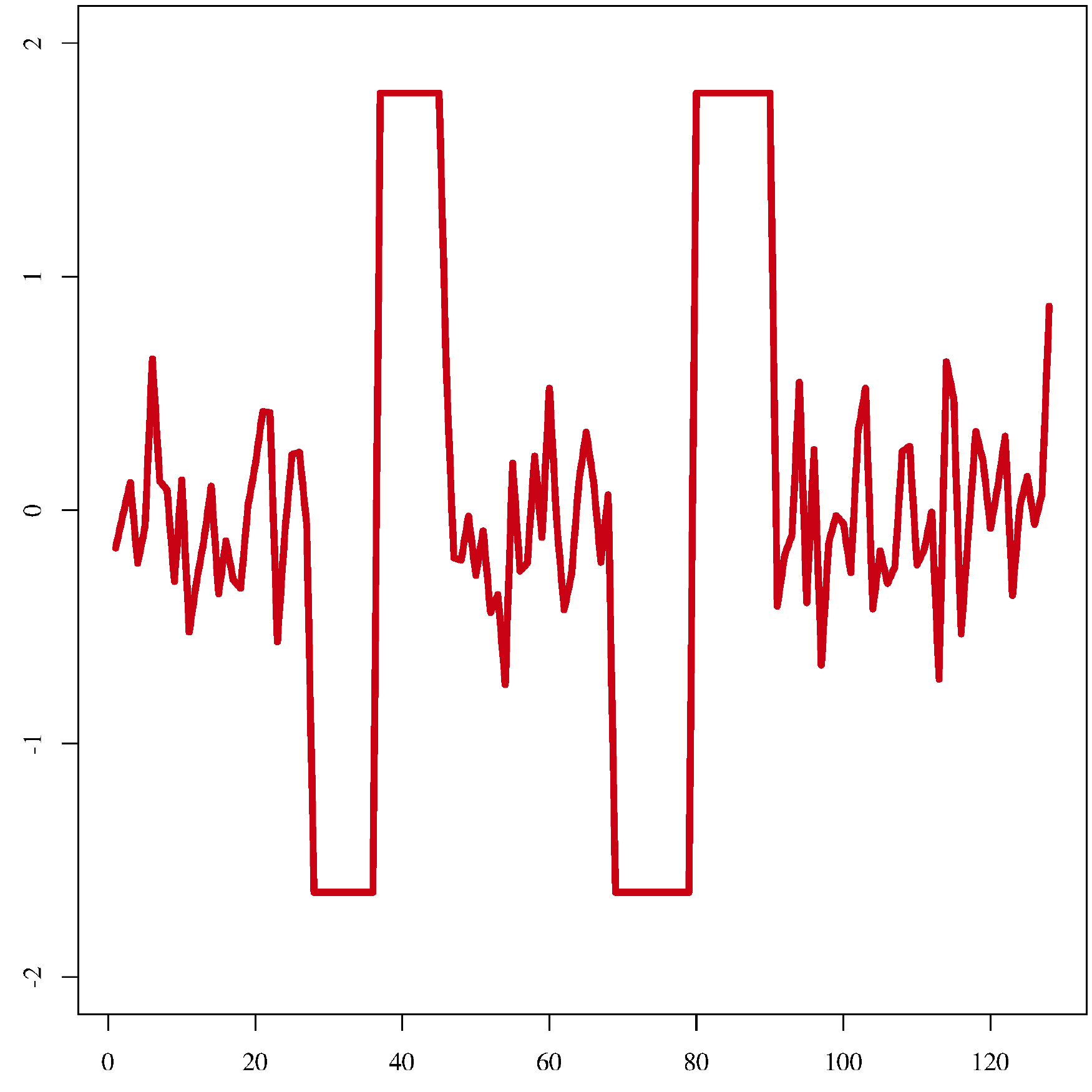}} 
\subfloat[]{\label{fig:two_patterns_ds_2} \includegraphics[scale=0.16]{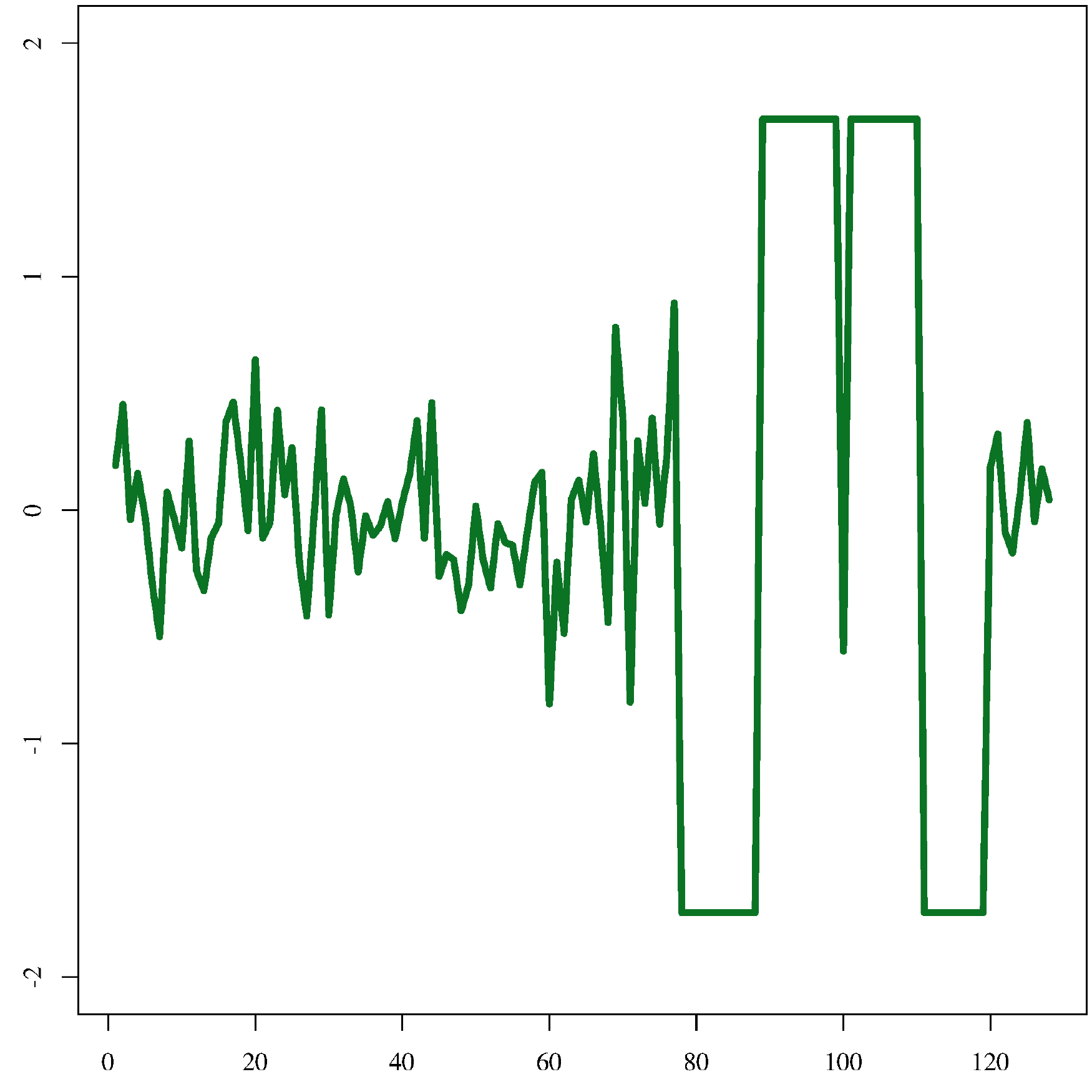}}\\
\subfloat[]{\label{fig:two_patterns_ds_3} \includegraphics[scale=0.16]{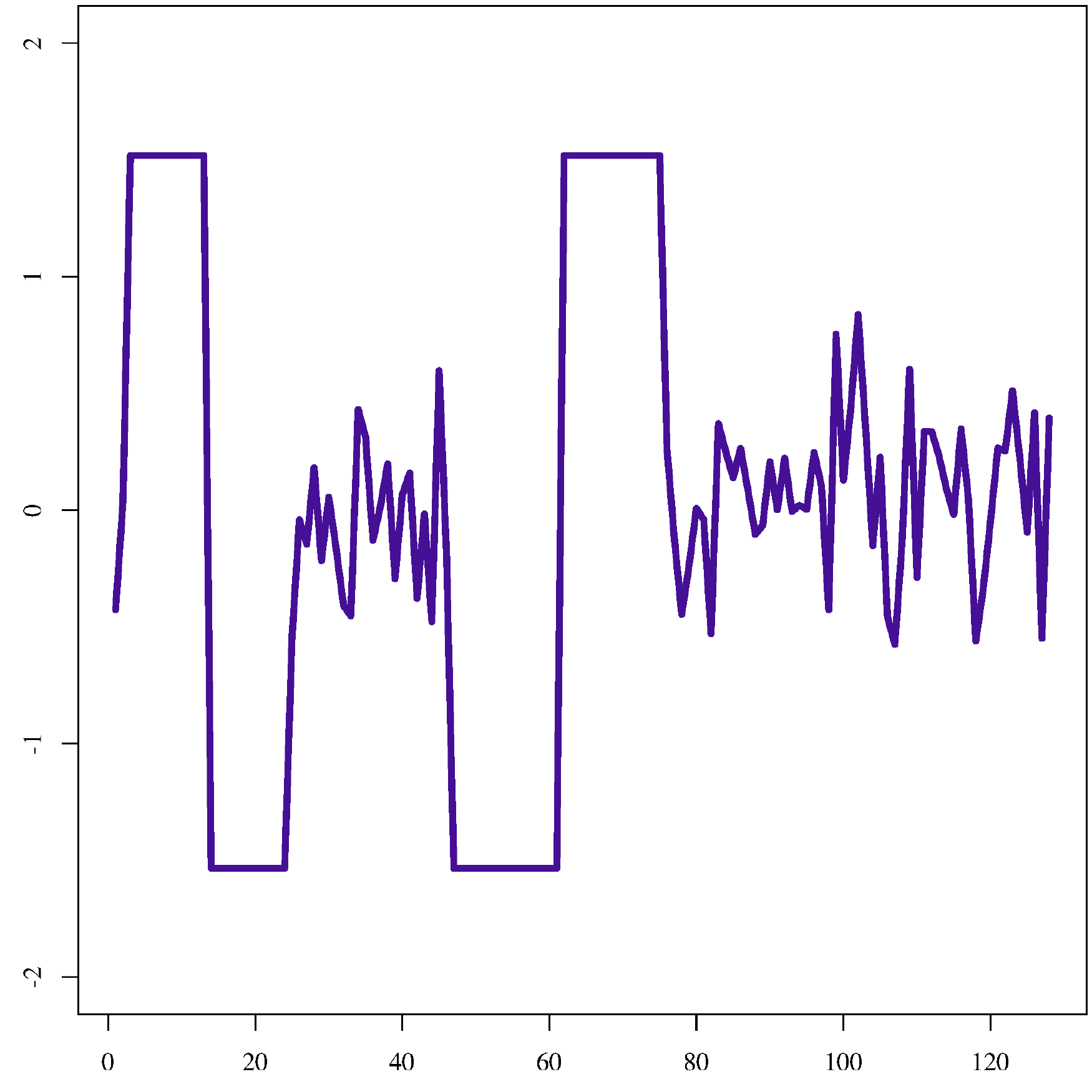}}
\subfloat[]{\label{fig:two_patterns_ds_4} \includegraphics[scale=0.16]{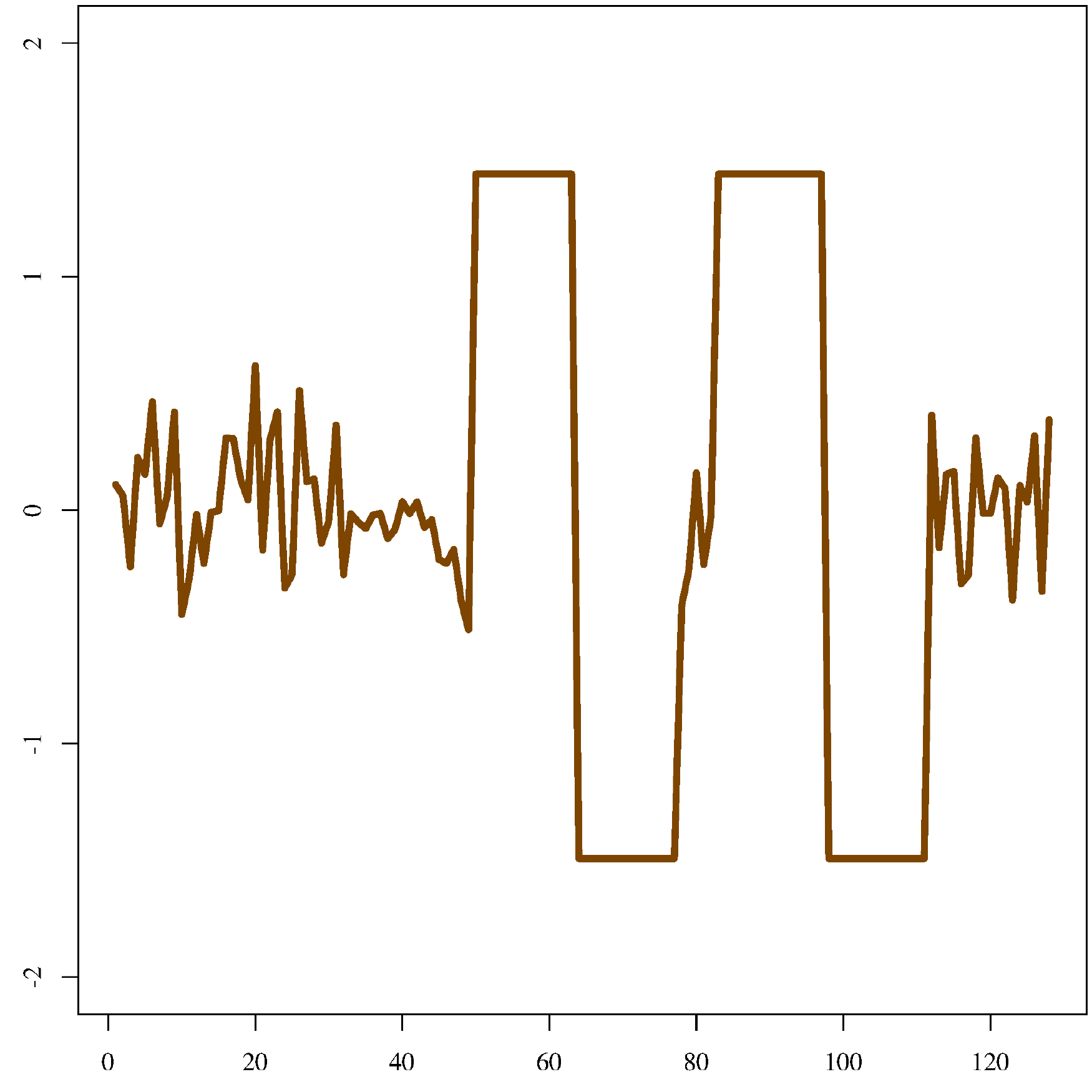}}
\caption{The two patterns data set is composed by the sequence of two patterns: an upward (U) step, which goes from $-5$ to 5, and a downward (D) step, which goes from 5 to $-5$. The order which these patterns occur define each group: \protect\subref{fig:two_patterns_ds_1} UU, \protect\subref{fig:two_patterns_ds_2} UD, \protect\subref{fig:two_patterns_ds_3} DU and \protect\subref{fig:two_patterns_ds_4} DD.}
\label{fig:two_patterns_ds} 
\end{figure}

In some cases, the similarity of time series is defined by repeating patterns that should be efficiently detected by clustering algorithms. We exemplify the efficiency of our method to detect different shape patterns in time series considering the two patterns data set \cite{geurts02}. It is composed by 1000 time series of length 128 divided into four groups. These groups are characterized by the occurrence of two different patterns in a defined order: an upward step (which goes from -5 to 5) and a downward step (which goes from 5 to -5). Using these two patterns, it is possible to define 4 groups: UU, UD, DU and DD. The group UU is defined by two upward steps, UD is defined by an upward step followed by a downward step, and the same logic defines DD and DU groups. According to these definition, clustering algorithms should be capable of detecting the order of patterns to correctly distinguish UD and DU. To make the problem harder, the position and duration of the patterns are randomized in such a way that there is no overlap. Around patterns, the series is characterized by an independent Gaussian noise. Figure \ref{fig:two_patterns_ds} illustrates the 4 groups of the data set.

\begin{figure}[!b]
  \centering
  \includegraphics[width=.37 \textwidth]{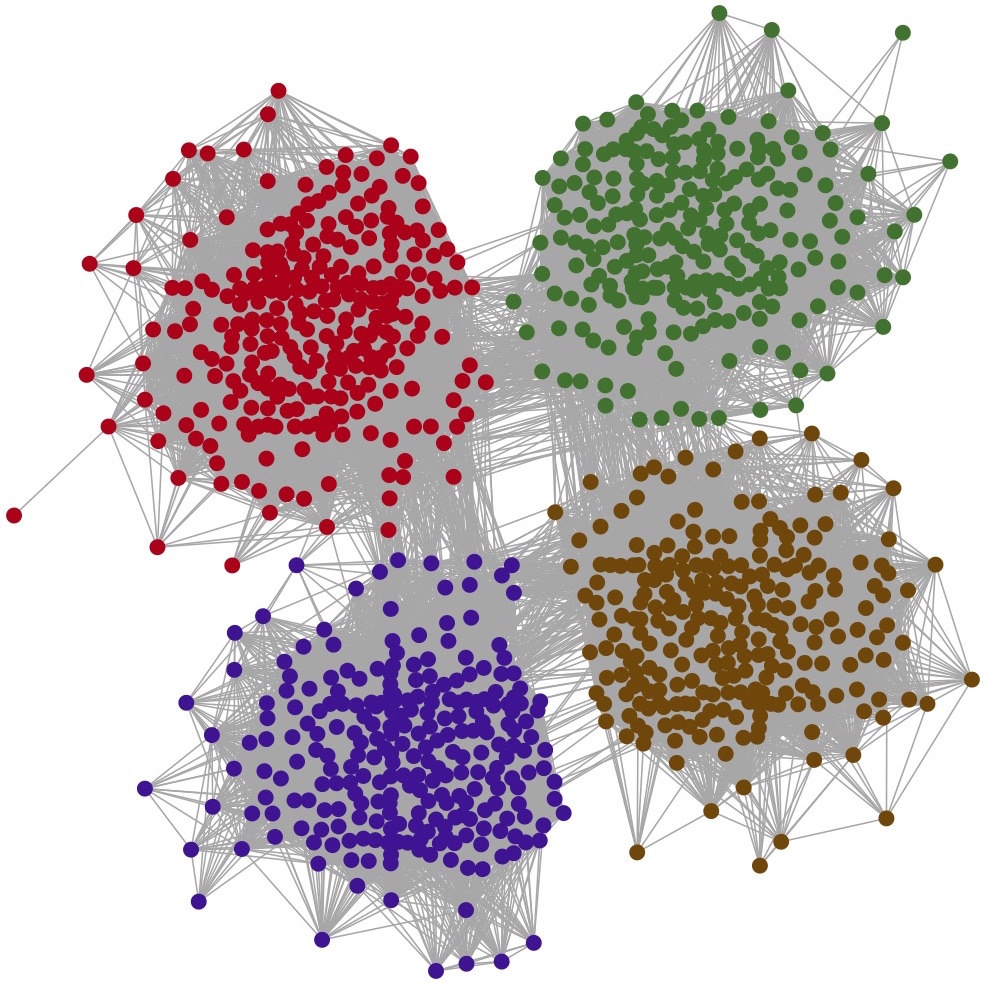}
  \caption{Network representation of the two patterns data set using the $\varepsilon$-NN construction method ($\varepsilon=44.91$) with DTW. Vertices colors indicate the 4 communities that represent each group of time series illustrated in Fig. \ref{fig:two_patterns_ds}. All the 1000 time series were correctly clustered ($RI=1$).} 
  \label{fig:two_patterns_net}
\end{figure}

Using the $\varepsilon$-NN construction method ($\varepsilon=44.91$) with DTW, it is possible to construct a network as shown in Fig. \ref{fig:two_patterns_net}, which represents the two patterns data set. After applying the multilevel community detection algorithm to this network, we get 4 communities, representing each group of time series. All the 1000 time series are correctly clustered ($RI=1$). The rival method (Tab. \ref{tab:rand_clustering}) that achieves the best clustering result for this data set is the single linkage with DTW: $RI=0.97$.

% ==============================================================================
% Section -
% ==============================================================================
\section{Conclusion}
\label{sec:final}

In this paper we present benefits of using community detection algorithms to perform time series clustering. According to the experimental results, we conclude that the best results are achieved using the $\varepsilon$-NN construction method with the DTW distance function and the multilevel community detection algorithm among the combinations under study. We have observed that intermediate values of $k$ and $\varepsilon$ lead to better clustering results (Sec. \ref{subsec:network_construction_influence}). 

For a fair comparison, we have also verified which distance function works better with each of the rival algorithms (Tab. \ref{tab:rand_clustering}). We compare those algorithms to our method using different data sets and we confirm that our method outperformed in most of the tested datasets. We have observed that our method has ability to detect groups of series even presenting time shifts and amplitude variations. All the facts indicate that using community detection algorithms for time series data clustering is an interesting approach.

Another advantage of the proposed approach is that it can be easily fit to specific clustering problems by changing the network construction method, the distance function or the community detection algorithm. Another advantage is that general improvements on these subroutines are applicable to our method.  

The proposed method has been developed considering only on univariate time series. However, the same idea can be extended to multivariate time series clustering at least in the following ways: 1) changing the time series distance function. In this case, we just need to use a new distance function designed for multivariate time series. The network construction method and the clustering method remain the same. 2) Changing the clustering method. In this case, a new clustering method has to be developed to deal with every series variables. One possible way is to apply our method to each variable and then use some criteria to merge the clustering results. As a future work, we plan to address this problem.

In this paper, we have made statistical comparisons of clustering accuracy based on the rand index. Although it is a good measure and presents good results, it would be interesting to evaluate the simulation results using different indexes. Another point is that we have compared the best rand indexes searching from a variation of $k$ and $\varepsilon$. In many real datasets, it would be infeasible to do such a searching due to the time consuming. As future works, we plan to propose automatic strategies for choosing the best number of neighbors ($k$ and $\varepsilon$) and speeding up the network construction method, instead of using the naive method. We also plan to apply the idea to solve other kinds of problems in time series analysis, such as time series prediction. 

\section*{Acknowledgments}

We would like to thank CNPq, CAPES and FAPESP for supporting this research. We thank the University of S\~ao Paulo for providing the computational infrastructure of the cloud computing that allowed the experiments. We would like to thank Prof. Eamonn Keogh for providing the datasets \cite{ucr14}. We also want to thank the developers from \texttt{igraph} \cite{igraph06}, \texttt{TSdist} \cite{tsdist14} and \texttt{TSclust} \cite{tsclust14} R libraries for making easier the development of this paper.



 \bibliographystyle{elsart-num-sort} 
 \bibliography{ref}

\begin{thebibliography}{10}
\expandafter\ifx\csname url\endcsname\relax
  \def\url#1{\texttt{#1}}\fi
\expandafter\ifx\csname urlprefix\endcsname\relax\def\urlprefix{URL }\fi

\bibitem{batista14}
G.~E. A. P.~A. Batista, E.~J. Keogh, O.~M. Tataw, V.~M.~A. de~Souza, Cid: an
  efficient complexity-invariant distance for time series, Data Mining and
  Knowledge Discovery 28~(3) (2014) 634--669.

\bibitem{berndt94}
D.~J. Berndt, J.~Clifford, {Using Dynamic Time Warping to Find Patterns in Time
  Series}, in: KDD Workshop, 1994, pp. 359--370.

\bibitem{blondel08}
V.~D. Blondel, J.-L. Guillaume, R.~Lambiotte, E.~Lefebvre, Fast unfolding of
  communities in large networks, Journal of Statistical Mechanics: Theory and
  Experiment 2008~(10) (2008) P10008.

\bibitem{boccaletti06}
S.~Boccaletti, V.~Latora, Y.~Moreno, M.~Chavez, D.-U. Hwang, Complex networks:
  Structure and dynamics, Physics Reports 424~(4--5) (2006) 175 -- 308.

\bibitem{brandmaier11}
A.~M. Brandmaier, Permutation distribution clustering and structural equation
  model trees, Ph.D. thesis, Universit\"{a}t des Saarlandes (2011).

\bibitem{chappelier96}
J.~Chappelier, A.~Grumbach, A kohonen map for temporal sequences, in: In
  Proceedings of the Conference on Neural Networks and Their Applications,
  1996, pp. 104--110.

\bibitem{chen09}
J.~Chen, H.-r. Fang, Y.~Saad, Fast approximate knn graph construction for high
  dimensional data via recursive lanczos bisection, J. Mach. Learn. Res. 10
  (2009) 1989--2012.

\bibitem{clauset04}
A.~Clauset, M.~E.~J. Newman, C.~Moore, Finding community structure in very
  large networks, Phys. Rev. E 70 (2004) 066111.

\bibitem{igraph06}
G.~Csardi, T.~Nepusz, The igraph software package for complex network research,
  InterJournal Complex Systems (2006) 1695.

\bibitem{casado03}
D.~C. de~Lucas, Classification techniques for time series and functional data,
  Ph.D. thesis, Universidad Carlos III de Madrid (2003).

\bibitem{demsar06}
J.~Dem\v{s}ar, Statistical comparisons of classifiers over multiple data sets,
  J. Mach. Learn. Res. 7 (2006) 1--30.

\bibitem{esling12}
P.~Esling, C.~Agon, {Time-series data mining}, ACM Computing Surveys 45~(1)
  (2012) 1--34.

\bibitem{extra15}
L.~N. Ferreira, L.~Zhao, Code and extra information for the paper: Time series
  clustering via community detection in networks,
  http://lnferreira.github.io/time\_series\_clustering\_via\_community\_detection,
  accessed Fev-2015 (Fev 2015).

\bibitem{fortunato10}
S.~Fortunato, Community detection in graphs, Physics Reports 486~(3--5) (2010)
  75--174.

\bibitem{frentzos07}
E.~Frentzos, K.~Gratsias, Y.~Theodoridis, Index-based most similar trajectory
  search, in: Data Engineering, 2007. ICDE 2007. IEEE 23rd International
  Conference on, 2007, pp. 816--825.

\bibitem{gan07}
G.~Gan, C.~Ma, J.~Wu, Data Clustering: Theory, Algorithms, and Applications,
  Society for Industrial and Applied Mathematics, 2007.

\bibitem{geurts02}
P.~Geurts, Contributions to decision tree induction: Bias/variance tradeoff and
  time series classification, Ph.D. thesis, Department of Electrical
  Engineering, University of Liege, Belgium (2002).

\bibitem{newman02}
M.~Girvan, M.~E.~J. Newman, Community structure in social and biological
  networks, Proceedings of the National Academy of Sciences 99~(12) (2002)
  7821--7826.

\bibitem{golay98}
X.~Golay, S.~Kollias, G.~Stoll, D.~Meier, A.~Valavanis, P.~Boesiger, A new
  correlation-based fuzzy logic clustering algorithm for fmri, Magnetic
  Resonance in Medicine 40~(2) (1998) 249--260.

\bibitem{guo08}
C.~Guo, H.~Jia, N.~Zhang, Time series clustering based on ica for stock data
  analysis, in: Wireless Communications, Networking and Mobile Computing, 2008.
  WiCOM '08. 4th International Conference on, 2008, pp. 1--4.

\bibitem{halkidi01}
M.~Halkidi, Y.~Batistakis, M.~Vazirgiannis, On clustering validation
  techniques, J. Intell. Inf. Syst. 17~(2-3) (2001) 107--145.

\bibitem{keogh04}
E.~Keogh, S.~Lonardi, C.~A. Ratanamahatana, Towards parameter-free data mining,
  in: Proceedings of the Tenth ACM SIGKDD International Conference on Knowledge
  Discovery and Data Mining, KDD '04, ACM, New York, NY, USA, 2004, pp.
  206--215.

\bibitem{ucr14}
E.~Keogh, Q.~Zhu, B.~Hu, Y.~Hao., X.~Xi, L.~Wei, C.~A. Ratanamahatana, {The UCR
  time series dataset}, \url{http://www.cs.ucr.edu/~eamonn/time_series_data/},
  [Online; accessed Sep-2014] (2008).

\bibitem{maharaj00}
E.~Maharaj, Cluster of time series, Journal of Classification 17~(2) (2000)
  297--314.

\bibitem{tsclust14}
P.~M. Manso, J.~A. Vilar, TSclust: Time series clustering utilities, r package
  version 1.2.1 (2014).

\bibitem{moller03}
C.~M{\"o}ller-Levet, F.~Klawonn, K.-H. Cho, O.~Wolkenhauer, Fuzzy clustering of
  short time-series and unevenly distributed sampling points, in: Advances in
  Intelligent Data Analysis V, vol. 2810 of Lecture Notes in Computer Science,
  Springer Berlin Heidelberg, 2003, pp. 330--340.

\bibitem{tsdist14}
U.~Mori, A.~Mendiburu, J.~Lozano, TSdist: Distance Measures for Time Series
  data., r package version 1.2 (2014).

\bibitem{pons05}
P.~Pons, M.~Latapy, Computing communities in large networks using random walks,
  in: Computer and Information Sciences - ISCIS 2005, vol. 3733 of Lecture
  Notes in Computer Science, Springer Berlin Heidelberg, 2005, pp. 284--293.

\bibitem{raghavan07}
U.~N. Raghavan, R.~Albert, S.~Kumara, Near linear time algorithm to detect
  community structures in large-scale networks, Phys. Rev. E 76 (2007) 036106.

\bibitem{rosvall08}
M.~Rosvall, C.~T. Bergstrom, Maps of random walks on complex networks reveal
  community structure, Proceedings of the National Academy of Sciences 105~(4)
  (2008) 1118--1123.

\bibitem{SilvaZhao2012}
T.~C. Silva, L.~Zhao, Stochastic competitive learning in complex networks, IEEE
  Trans. Neural Networks and Learning Systems 23 (2012) 385--397.

\bibitem{smyth97}
P.~Smyth, Clustering sequences with hidden markov models, in: Advances in
  Neural Information Processing Systems, MIT Press, 1997, pp. 648--654.

\bibitem{vlachos02}
M.~Vlachos, G.~Kollios, D.~Gunopulos, Discovering similar multidimensional
  trajectories, in: Data Engineering, 2002. Proceedings. 18th International
  Conference on, 2002, pp. 673--684.

\bibitem{keogh13}
X.~Wang, A.~Mueen, H.~Ding, G.~Trajcevski, P.~Scheuermann, E.~Keogh,
  Experimental comparison of representation methods and distance measures for
  time series data, Data Mining and Knowledge Discovery 26~(2) (2013) 275--309.

\bibitem{liao05}
T.~Warren~Liao, Clustering of time series data-a survey, Pattern Recogn.
  38~(11) (2005) 1857--1874.

\bibitem{yimin02}
Y.~Xiong, D.-Y. Yeung, Mixtures of arma models for model-based time series
  clustering, in: Data Mining, 2002. ICDM 2003. Proceedings. 2002 IEEE
  International Conference on, 2002, pp. 717--720.

\bibitem{xiong04}
Y.~Xiong, D.-Y. Yeung, Time series clustering with arma mixtures, Pattern
  Recognition 37~(8) (2004) 1675 -- 1689.

\bibitem{yi00}
B.-K. Yi, C.~Faloutsos, Fast time sequence indexing for arbitrary lp norms, in:
  Proceedings of the 26th International Conference on Very Large Data Bases,
  VLDB '00, Morgan Kaufmann Publishers Inc., San Francisco, CA, USA, 2000, pp.
  385--394.

\bibitem{zakaria12}
J.~Zakaria, A.~Mueen, E.~Keogh, Clustering time series using
  unsupervised-shapelets, in: Proceedings of the 2012 IEEE 12th International
  Conference on Data Mining, ICDM '12, IEEE Computer Society, Washington, DC,
  USA, 2012, pp. 785--794.

\bibitem{zhang06}
H.~Zhang, T.~B. Ho, Y.~Zhang, M.~S. Lin, Unsupervised feature extraction for
  time series clustering using orthogonal wavelet transform, Informatica
  (Slovenia) 30~(3) (2006) 305--319.

\bibitem{zhang11}
X.~Zhang, J.~Liu, Y.~Du, T.~Lv, A novel clustering method on time series data,
  Expert Syst. Appl. 38~(9) (2011) 11891--11900.

\end{thebibliography}





% \end{thebibliography}

% \appendix[Data Set Description]
% or
\appendix  % for no appendix heading


% \appendices
\section{Data Set Description}
\label{append:data_set}

In the simulations of this paper, we have used 45 time series data sets taken from the UCR repository \cite{ucr14}. This repository is composed of real and synthetic data sets divided in training and test sets. For our experiments, we consider only the training set and the test sets are discarded. These datasets have been generated by various authors and donated to the UCR repository. The labels of each dataset are not defined by the UCR, but are defined by the authors themselves according to the specific dataset domain. Therefore, we have to assume that the labels are correct. Table \ref{tab:datasets} describes each data set used in this paper.

\singlespacing
\begin{table}[ht]
\centering
\scriptsize
\begin{threeparttable}
\caption{The UCR time series data sets used in the experiments}
\label{tab:datasets}
% \rowcolors{2}{gray!25}{white}
\begin{tabular}{lcccc}
  \toprule
  Data set                        & Num.       & Time series & Num.           \\
                                  & objects    & length & classes        \\
  \midrule
   Adiac & 390 & 176 &  37 \\
   Beef &  30 & 470 &   5 \\
   Car &  60 & 577 &   4 \\
  CBF &  30 & 128 &   3 \\
   ChlorineConcentration & 467 & 166 &   3 \\
   CinC\_ECG\_torso &  40 & 1639 &   4 \\
   Coffee &  28 & 286 &   2 \\
   Cricket\_X & 390 & 300 &  12 \\
   Cricket\_Y & 390 & 300 &  12 \\
   Cricket\_Z & 390 & 300 &  12 \\
   DiatomSizeReduction &  16 & 345 &   4 \\
   ECG & 100 &  96 &   2 \\
   ECGFiveDays &  23 & 136 &   2 \\
  Face (all) & 560 & 131 &  14 \\
   Face (four) &  24 & 350 &   4 \\
   FacesUCR & 200 & 131 &  14 \\
   Fish & 175 & 463 &   7 \\
  Gun-Point &  50 & 150 &   2 \\
   Haptics & 155 & 1092 &   5 \\
   InlineSkate & 100 & 1882 &   7 \\
   ItalyPowerDemand &  67 &  24 &   2 \\
   Lightning-2 &  60 & 637 &   2 \\
   Lightning-7 &  70 & 319 &   7 \\
   MALLAT &  55 & 1024 &   8 \\
   MedicalImages & 381 &  99 &  10 \\
   MoteStrain &  20 &  84 &   2 \\
   % Non-Invasive Fetal ECG Thorax1 & 1800 & 750 &  42 \\
   % Non-Invasive Fetal ECG Thorax2 & 1800 & 750 &  42 \\
   OliveOil &  30 & 570 &   4 \\
  OSU Leaf & 200 & 427 &   6 \\
   Plane & 105 & 144 &   7 \\
   SonyAIBORobot Surface &  20 &  70 &   2 \\
   SonyAIBORobot SurfaceII &  27 &  65 &   2 \\
   StarLightCurves & 1000 & 1024 &   3 \\
  Swedish Leaf & 500 & 128 &  15 \\
   Symbols &  25 & 398 &   6 \\
  Synthetic Control & 300 &  60 &   6 \\
  Trace & 100 & 275 &   4 \\
  Two Patterns & 1000 & 128 &   4 \\
   TwoLeadECG &  23 &  82 &   2 \\
   uWaveGestureLibrary\_X & 896 & 315 &   8 \\
   uWaveGestureLibrary\_Y & 896 & 315 &   8 \\
   uWaveGestureLibrary\_Z & 896 & 315 &   8 \\
   Wafer & 1000 & 152 &   2 \\
   WordsSynonyms & 267 & 270 &  25 \\
   Words 50 & 450 & 270 &  50 \\
   Yoga & 300 & 426 &   2 \\
  \bottomrule
\end{tabular}
\end{threeparttable}
\end{table}
\doublespacing

\end{document}